\documentclass[11pt]{article}

\usepackage[final]{acl}

\usepackage{times}
\usepackage{latexsym}

\usepackage[T1]{fontenc}

\usepackage[utf8]{inputenc}

\usepackage{microtype}

\usepackage{inconsolata}

\usepackage{graphicx}
\usepackage{amsmath}
\usepackage{url}

\usepackage{booktabs}   
\usepackage{xcolor}     
\usepackage{xspace}     

\usepackage{times}
\usepackage{latexsym}
\usepackage{enumitem}
\usepackage{listings}
\usepackage{url}
\usepackage[T1]{fontenc}

\usepackage[utf8]{inputenc}

\usepackage{microtype}

\usepackage{inconsolata}

\usepackage{longtable}
\usepackage{graphicx}

\usepackage{caption}
\usepackage{subcaption}

\usepackage{colortbl}
\usepackage{makecell}
\usepackage{multirow}
\usepackage{supertabular}
\usepackage{placeins}

\usepackage{subfiles}
\usepackage{enumitem}

\usepackage{algorithm}
\usepackage{algpseudocode}
\usepackage{amsmath}

\usepackage{cleveref}

\setlist[itemize]{leftmargin=1em}  

\usepackage{booktabs}

\usepackage{tcolorbox}
\newcounter{promptno}[section]
\newlength\mystoreparindent
\newenvironment{prompt}[1][]
{
  \setlength{\mystoreparindent}{\the\parindent}
  \setlength{\parindent}{0pt}
  \refstepcounter{promptno}
  \par\medskip
  \noindent
  \begin{tcolorbox}[left=1pt,right=1pt]
  \textsc{{template \small\thesubsection.\thepromptno}}\\
  \small
  \tt
}{
  \end{tcolorbox}
  \setlength{\parindent}{\mystoreparindent}
  \medskip
}

\usepackage{amssymb}

\DeclareUnicodeCharacter{25A2}{$\square$}
\usepackage{color}

\newcolumntype{T}{>{\ttfamily}l}

\tcbuselibrary{skins}

\definecolor{a-colour}{RGB}{135,74,175}
\definecolor{b-colour}{RGB}{243,145,9}
\definecolor{gm-colour}{RGB}{128,128,128}
\definecolor{wordlegreen}{rgb}{0.4,0.79,0.17}
\definecolor{wordlered}{rgb}{0.86,0.31,0.29}
\definecolor{wordleyellow}{rgb}{1,1,0}

\def \dcolwidth {3cm}
\def \skipcolwidth {0.8cm}
\def \arccorner {3mm}
\def \disttocorner {6pt}

\newtcolorbox[use counter=nbubbles]{a-gm}[1]{
    colback=a-colour!10!white,
    colframe=a-colour,
    fonttitle=\bfseries\tiny,
    fontupper=\footnotesize,
    title={#1},
    sharp corners=west,
    arc=\arccorner,
    width=\dcolwidth,
    left skip=0cm,
    top=0pt,
    bottom=0pt,
    left=0pt,
    right=\disttocorner,
    before={\vspace{-0.1cm}},
    boxrule=0.5pt,
    enhanced,
    attach boxed title to top left={yshift=-0.1mm},
    boxed title style={size=small,colback=a-colour},
    overlay unbroken and first = {
    \node[text width=0.4cm,draw=black,line width=0.1mm,align=center,gray] at (-0.5,0.2) {\footnotesize \thetcbcounter};
  }
}

\newtcolorbox[use counter=nbubbles]{gm-a}[1]{
    colback=gm-colour!5!white,
    colframe=gm-colour,
    fonttitle=\bfseries\tiny,
    fontupper=\footnotesize,
    title={#1},
    sharp corners=east,
    arc=\arccorner,
    width=\dcolwidth + \skipcolwidth,
    left skip=\skipcolwidth,
    top=0pt,
    bottom=0pt,
    left=\disttocorner,
    right=0pt,
    before={\vspace{-0.1cm}},
    boxrule=0.5pt,
    halign title=flush right,
    enhanced,
    attach boxed title to top right={yshift=-0.1mm},
    boxed title style={size=small,colback=gm-colour},
    overlay unbroken and first = {
    \node[anchor=north east,text width=0.4cm,draw=black,line width=0.1mm,align=center,gray] at (-0.9,0.55) {\footnotesize\thetcbcounter};}
}

\newtcolorbox[use counter=nbubbles]{b-gm}[1]{
    colback=b-colour!10!white,
    colframe=b-colour,
    fonttitle=\bfseries\tiny,
    fontupper=\footnotesize,
    title={#1},
    sharp corners=east,
    arc=\arccorner,
    width=\dcolwidth + \skipcolwidth + \dcolwidth +  \skipcolwidth,
    left skip=\dcolwidth + \skipcolwidth + \skipcolwidth,
    top=0pt,
    bottom=0pt,
    left=\disttocorner,
    right=0pt,
    before={\vspace{-0.1cm}},
    boxrule=0.5pt,
    halign title=flush right,
    enhanced,
    attach boxed title to top right={yshift=-0.1mm},
    boxed title style={size=small,colback=b-colour},
    overlay unbroken and first = {
    \node[anchor=north east,text width=0.4cm,draw=black,line width=0.1mm,align=center,gray] at (-4.7,0.55) {\footnotesize\thetcbcounter};}
}

\newtcolorbox[use counter=nbubbles]{gm-b}[1]{
    colback=gm-colour!5!white,
    colframe=gm-colour,
    fonttitle=\bfseries\tiny,
    fontupper=\footnotesize,
    title={#1},
    sharp corners=west,
    arc=\arccorner,
    width=\dcolwidth + \skipcolwidth + \dcolwidth,
    left skip=\dcolwidth + \skipcolwidth,
    top=0pt,
    bottom=0pt,
    left=0pt,
    right=\disttocorner,
    before={\vspace{-0.1cm}},
    boxrule=0.5pt,
    enhanced,
    attach boxed title to top left={yshift=-0.1mm},
    boxed title style={size=small,colback=gm-colour},
    overlay unbroken and first = {
    \node[anchor=north east,text width=0.4cm,draw=black,line width=0.1mm,align=center,gray] at (-3.9,0.55) {\footnotesize\thetcbcounter};}
}

\newtcolorbox[use counter=nbubbles]{gm-gm}[1]{
    colback=gm-colour!5!white,
    colframe=gm-colour,
    fonttitle=\bfseries\tiny,
    fontupper=\footnotesize,
    title={#1},
    sharp corners,
    width=\dcolwidth + \dcolwidth + \skipcolwidth + \skipcolwidth,
    leftright skip=\dcolwidth,
    top=0pt,
    bottom=0pt,
    left=0pt,
    right=0pt,
    before={\vspace{-0.1cm}},
    boxrule=0.5pt,
    halign title=center,
    enhanced,
    attach boxed title to top center={yshift=-0.1mm},
    boxed title style={size=small,colback=gm-colour},
    overlay unbroken and first = {
    \node[anchor=north east,text width=0.4cm,draw=black,line width=0.1mm,align=center,gray] at (-3.1,0.55) {\footnotesize\thetcbcounter};}
}

\title{The Image Reconstruction Game: Drawing Common\\
Ground Through Iterative Multimodal Dialogue}

\author{%
Sherzod Hakimov${^\mathbf{1}}$, 
Mattia \text{D'Agostini}${^\mathbf{1}}$,
Ivan Samodelkin${^\mathbf{1}}$,
\textbf{David Schlangen}${^\mathbf{1,2}}$\\$^{\mathbf{1}}$Computational Linguistics, Department of Linguistics\\
University of Potsdam, Germany\\
$^{\mathbf{2}}$German Research Center for Artificial Intelligence (DFKI), Berlin, Germany\\
{\texttt{\{firstname.lastname\}@uni-potsdam.de}}
}

\begin{document}
\maketitle

\begin{abstract}
We introduce the Image Reconstruction Game, a fully automated benchmark in which a vision-language model issues corrective instructions to an image generator across multiple turns, making accumulated common ground directly observable as a rendered image.
Benchmarking two Describer models crossed with two Generator models across seven image categories, we find that the describer is the dominant factor in reconstruction quality, while the generator determines whether iterative refinement helps or hurts. Mathematical and geometric images pose the greatest challenge. The describer's token budget strongly affects convergence: shorter budgets yield sparser first renderings with more room for visible improvement, while longer budgets raise absolute quality but leave less to fix. Stronger describers use a richer correction vocabulary spanning spatial, numeric, and structural categories, while weaker describers concentrate on surface properties and tend to stop after a few turns. Human validation shows that the best automated judge reaches only slight-to-fair agreement with human preferences, and automated scores require human recalibration to be used reliably.
\end{abstract}

\section{Introduction}
\label{sec:introduction}

\begin{figure}[t]
    \centering
    \includegraphics[width=1\linewidth]{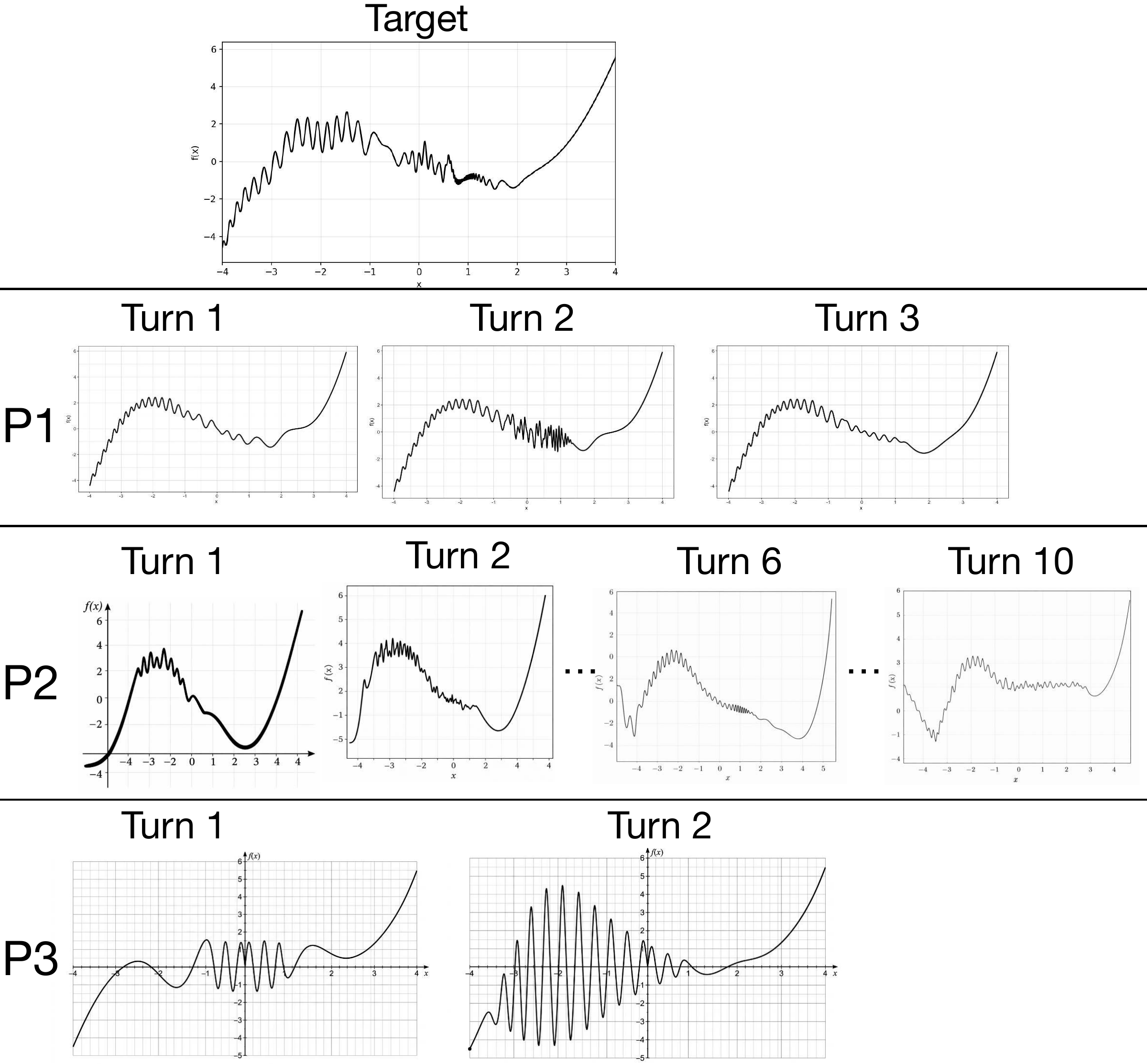}
\caption{In the IR game, one model (the describer) tries to steer another (the generator) towards faithful reproduction of a target, using only natural language descriptions, over several turns. Figure shows example of how the generated image develops in response to the instructions (which are not shown), for three different model pairings and the same target.} 
    \label{fig:sample_renderings}
    \vspace*{-.5cm}
\end{figure}

\begin{figure*}[t]
\begin{center}
  \includegraphics[width=.9\linewidth]{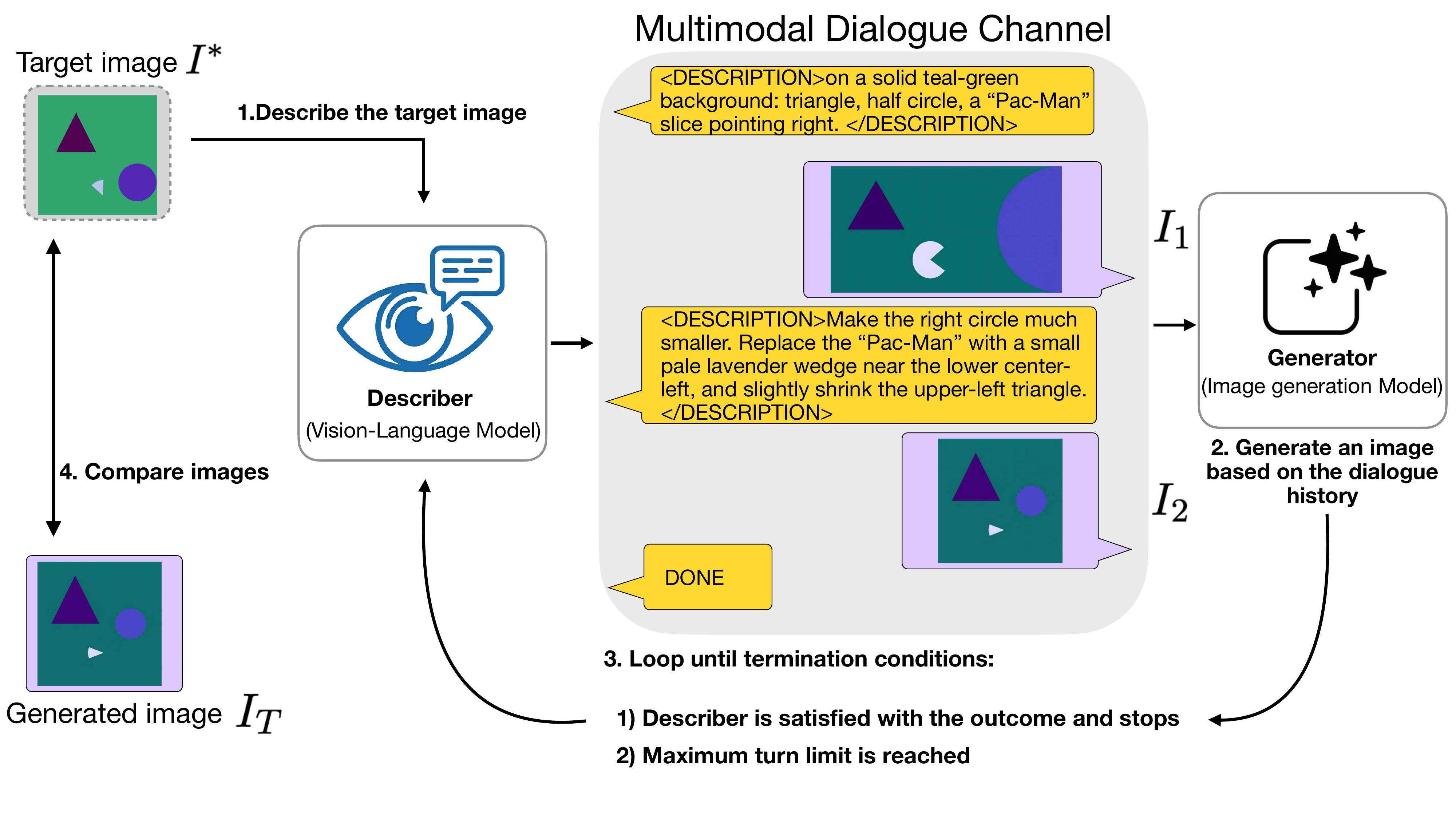}
\end{center}
  \vspace*{-.5cm}  
  \caption{%
    Overview of the Image Reconstruction Game.
    The \emph{Describer} (a VLM) has access to the target image~$I^*$ and
    communicates with the \emph{Generator} (an image model) exclusively through
    natural language.
    The Generator produces a rendering ($I_1$ or $I_2$) from the accumulated dialogue; the
    Describer observes both~$I^*$ and the rendering and issues a corrective
    instruction for the next turn.
    Each rendering makes the current state of common ground directly visible,
    and the loop continues until the Describer signals convergence or a turn
    limit is reached. Comparison is done between target image $I^*$ and the final turn generated image $I_T$.}
  \label{fig:overview}
  \vspace*{-.3cm}  
\end{figure*}

Imagine you have a picture ``in your head'' -- a scene that you saw, a diagram that you think could explain something, etc. -- that you want to see materialised. Image generation models aim to address this use case: You describe what is ``in your head'' using language, and the model generates an image. Where early models technically simply were functions from textual prompts to images, modern models set up this task in a more iterative, incremental and dialogic way, where earlier attempts can be corrected and modified. Using terminology from the study of conversations, this can be described as a process of \textit{conversational grounding} \cite{clark:ul}, where common ground between the interactants is reached through an interactive process of clarification and correction. This paper uses this setup to test current models on the various components of this task: How good are current language and vision models at describing an image in such a way as to steer an image generation model to reproduce it, and how good are such image generation models at being steered? Figure~\ref{fig:sample_renderings} illustrates the outputs of three model pairs across turns to reconstruct the target image through dialogue.

This adds a new dimension to the testing of models that combine linguistic and visual capabilities, as commonly the focus in their evaluation is on non-interactive tasks---captioning, VQA, image--text alignment---that do not test the capacities~\cite{ALFRED20}.
Settings such as those used in CoDraw \citep{kim2019codraw} establish that iterative
visual communication can be studied as a collaborative task with measurable
outcomes, but rely on human players and fixed symbolic representations.
Conversational image generation systems
\citep{el2019tell,huang2025dialoggen,ma2026talk2image} place the human in the
describer role, targeting user intent rather than evaluating the describer itself.
Neither line holds one component fixed while varying the other, and neither
connects communication trajectories to human perceptual judgements of outcomes.

We introduce the \textbf{Image Reconstruction Game}, a fully automated
multimodal dialogue game that fills these gaps (see Figure~\ref{fig:overview}).
A \emph{describer} model (a VLM) observes a target image and must steer
an \emph{image generator} to reproduce it through a sequence of natural-language
turns; at each turn the generator's rendering makes the current state of common
ground directly visible, and the describer observes both \textit{target} and all \textit{renderings up to that point in the dialogue} before issuing a corrective instruction.
No human is in the loop: the game is played entirely between models under
controlled, reproducible conditions.
This design lets us investigate four research questions:
\textbf{RQ1}~Can VLMs detect fine-grained perceptual differences between paired
images and verbalise them as actionable, corrective instructions?
\textbf{RQ2}~Do those instructions measurably reduce the difference between the
target and the current rendering over turns?
\textbf{RQ3}~Does the describer--generator pair build common ground
progressively, as reflected in both automated similarity scores and human
ratings?
\textbf{RQ4}~Which automated image-similarity metrics best track human
perceptual judgements in this setting?

Addressing this last question is non-trivial: embedding-based similarity metrics
are known to diverge substantially from human perceptual
judgements~\citep{DBLP:conf/iclr/GhazanfariAKKG24}, and LLM-based judges---while better
correlated with human ratings than CLIP or BLIP in text-to-image
settings~\citep{10.5555/3666122.3667123}---cannot be assumed to carry over to
image reconstruction without direct empirical
validation~\citep{DBLP:conf/eccv/FuHLFWLRSMK24,marjieh2024large}.
Our contributions are: (i)~a new task formulation and instance set spanning seven
image categories that vary the difficulty of perception and verbalisation;
(ii)~a benchmark comparing two describer models against two generators,
isolating describer quality from generator variation;
(iii)~a human similarity rating study on a stratified subset, together with an
analysis of which automated metrics are most reliable proxies;
and (iv)~a qualitative analysis of what kinds of differences are hard to
perceive or verbalise\footnote{\url{https://github.com/clp-research/image-reconstruction-game}}.

\section{Related Work}
\label{sec:related}

CoDraw \citep{kim2019codraw} establishes the asymmetric Teller--Drawer paradigm
for iterative visual reconstruction through multi-turn dialogue, and related
work shows how common ground accumulates across turns when agents describe or
identify shared images
\citep{haber2019photobook,clark2021iconary,madureira2023instruction,DBLP:conf/eacl/TestoniF24,han2017drawandtell}.
Game-based frameworks have also been adopted as evaluation instruments for
language and multimodal models: \citet{chalamalasetti2023clembench} systematise
this for text-based chat models, and \citet{hakimov2025using} extend it to
multimodal models with games probing reference resolution and image comparison
\citep{zhong2021silg,mohapatra2023conversational,li2024mug}.
Our game inherits this paradigm where players are VLM models as describer and generative image model as generator, making
common ground directly observable as a rendered image at every turn.

A broad family of systems generates or refines images through multi-turn natural
language \citep{ullah2022review,el2019tell,lee2021visual,wei2023dialogpaint,%
huang2024promptcharm,he2025tdri,ma2025dialogdraw,huang2025dialoggen,%
wang2025twinco,ma2026talk2image,hu2026talkphoto}.
In all cases a human user provides instructions; the research question is how to
build a better generative pipeline, not how to evaluate the instruction-giver.

Co-creative drawing agents assist human sketchers in open-ended creative tasks
\citep{davis2016cocreative,davis2017quantifying,fan2019collabdraw,huang2020scones,%
zhang2021storydrawer,ibarrola2024collaborative,davis2025ai}, while emergent
communication research asks whether agents can evolve functional visual protocols
through interaction \citep{fernando2020drawing,mihai2021learning,qiu2022emergent,%
vinker2025sketchagent}.
Both treat visual output as a communicative medium, but target artistic
expression or protocol emergence rather than the fidelity of reconstruction by
existing models.

Human evaluation of generated images typically takes the form of pairwise
preference judgements or direct similarity ratings, and is used as the
ground-truth reference against which automated metrics are calibrated.
Standard metrics such as SSIM, LPIPS, DINO, and CLIP are known to diverge from
human mid-level perceptual judgements~\citep{DBLP:conf/iclr/GhazanfariAKKG24}, and
multimodal LLMs similarly under-perform humans on explicit visual similarity
tasks~\citep{DBLP:conf/eccv/FuHLFWLRSMK24}.
LipSim~\citep{DBLP:conf/iclr/GhazanfariAKKG24} and
LLMScore~\citep{10.5555/3666122.3667123} address this by providing provably
robust perceptual similarity estimates and aligning metrics to human quality
ratings respectively, while
\citet{marjieh2024large,DBLP:conf/cogsci/MarjiehSSJ022} show that LLMs can
approximate human similarity ratings across perceptual domains, though with
systematic biases.
We extend this line to image reconstruction, using human ratings to validate
LLM judges that also participated as Describer models.

Recent benchmarks specifically targeting mathematical and geometric visual
perception---including MathOPEval~\citep{mathopeval}, GGBench~\citep{ggbench},
MATHGLANCE~\citep{mathglance}---consistently show that diagram-level
geometric perception is a critical bottleneck in current VLMs.
Where these benchmarks measure whether a model can \emph{read} a geometric
figure, the Image Reconstruction Game asks whether a model can
\emph{articulate residual differences} between two such figures---a harder,
relational task.
Our finding that \textbf{Functions} and \textbf{Geometry} categories pose
the greatest challenge aligns with and extends this body of evidence to
corrective, goal-directed communication.

\section{The Image Reconstruction Game}
\label{sec:game}

\subsection{Task definition}
\label{sec:task}

\paragraph{Game description.}

The Image Reconstruction Game is a two-player dialogue game implemented within
the \texttt{clem} framework for game-based model evaluation
\citep{chalamalasetti2023clembench,hakimov2025using}.
In this framework, a programmatic \textsc{GameMaster} instantiates each game
episode with a given target image, manages turn-by-turn prompting of the two
players, enforces the communication protocol, and records the episode for
scoring.

The two players occupy asymmetric roles with different information access and capabilities
(see Figure~\ref{fig:overview}).
The \emph{Describer} is a Vision-and-Language Model that has full access
to the target image~$I^*$.
Its goal is to produce natural-language utterances that guide the Generator
toward a faithful visual reproduction of~$I^*$.
The \emph{Generator} is an image generation model that never observes~$I^*$
directly; it can only perceive the accumulated dialogue history---the sequence
of Describer utterances---and must render an image conditioned
on that text.

The game unfolds over multiple turns: at each turn the Describer observes
both~$I^*$ and the Generator's previous rendering ~$I_{t-1}$ ($t$ stands for the turn index), then issues a corrective
instruction or signals convergence; the Generator produces a new image based on the dialogue and previous rendering.

This iterative structure operationalises the progressive building of
\emph{common ground} \citep{alikhani2020achieving}: each rendering~$I_t$ is a
visible projection of the shared state constructed so far, and each Describer
utterance is a grounding act that attempts to reduce the residual gap.
The asymmetric information design---where only the Describer knows~$I^*$---
mirrors the Teller--Drawer setup of CoDraw \citep{kim2019codraw} and multimodal dialogue games (referential, collaboration, spatial navigation) studied by \citet{hakimov2025using}, while extending them to
open-ended natural-language communication with modern generative image models.

\paragraph{Capability tested.}
The game is designed to probe two interleaved capabilities that standard VLM
benchmarks~\cite{Antol_2015_ICCV,yue2023mmmu} do not jointly assess.
First, the Describer must \emph{perceive} fine-grained differences between~$I^*$
and~$I_{t-1}$--- a visual discrimination task that is considerably harder than
captioning a single image and that demands sensitivity to spatial, metric, and
relational properties.
Second, it must \emph{verbalise} those differences as specific corrective instructions for the Generator to act on, which requires goal-directed
language production rather than generic scene description.
The Generator, in turn, must interpret and act on incremental instructions
while maintaining coherence with what has already been rendered.
Jointly, these demands test whether a VLM can function as a goal-directed
communicative agent in a closed-loop, multimodal setting.

\subsection{Prompts and communication protocol}
\label{sec:protocol}

Each player's messages must conform to the communication protocol so that the
\textsc{GameMaster} can validate, parse, and relay them between players.
An episode continues until one of three stopping criteria is met:
(1)~Player~A signals convergence by producing a \textsc{done} action;
(2)~the maximum turn limit of $T{=}10$ is reached; or (3)~a player violates the protocol.

\paragraph{Player A (Describer).}
Player~A is a VLM that receives the target image~$I^*$ and the most recent
Generator rendering~$I_{t-1}$ at every turn (on turn~1, only~$I^*$ is
available, as no rendering exists yet).
The \textsc{GameMaster} presents these images with a prompt that asks Player~A
either to describe~$I^*$ in sufficient detail for an image generator to
reproduce it (turn~1), or to identify specific remaining differences between
$I^*$ and~$I_{t-1}$ and issue a targeted corrective instruction (turns $t > 1$).
Player~A must wrap its output in \texttt{<DESCRIPTION>}~\ldots~\texttt{</DESCRIPTION>}
tags so that the \textsc{GameMaster} can parse and relay the message to Player~B;
a response consisting solely of \textsc{done} (without tags) signals
convergence and ends the episode.
A token budget of 200~tokens per turn is imposed to enforce concise, targeted
corrections rather than exhaustive re-descriptions.

The correction turns place a higher demand on visual reasoning than the initial
description: corrections must be \emph{relational}---specifying what is wrong
relative to the current rendering---should focus on what to improve rather than descriptions of the
image from scratch.

\paragraph{Player B (Generator).}
Player~B is an image generation model that receives two inputs at each turn:
the previous rendering~$I_{t-1}$ and the text message produced by
Player~A in the same turn.
It does not have access to~$I^*$ at any point; its only signal is the
accumulated dialogue history together with the visual context of its own
previous output.
The \textsc{GameMaster} also passes the full sequence of prior Player~A
utterances alongside~$I_{t-1}$, ensuring that compositional information
established in early turns is
not lost when later turns focus on local corrections.
Player~B's sole output is a new rendered image~$I_t$, which is then forwarded
to Player~A as the basis for the next turn.

All prompt templates are given in Appendix~\ref{app:prompts}.

\begin{figure*}
    \centering
    \includegraphics[width=0.8\linewidth]{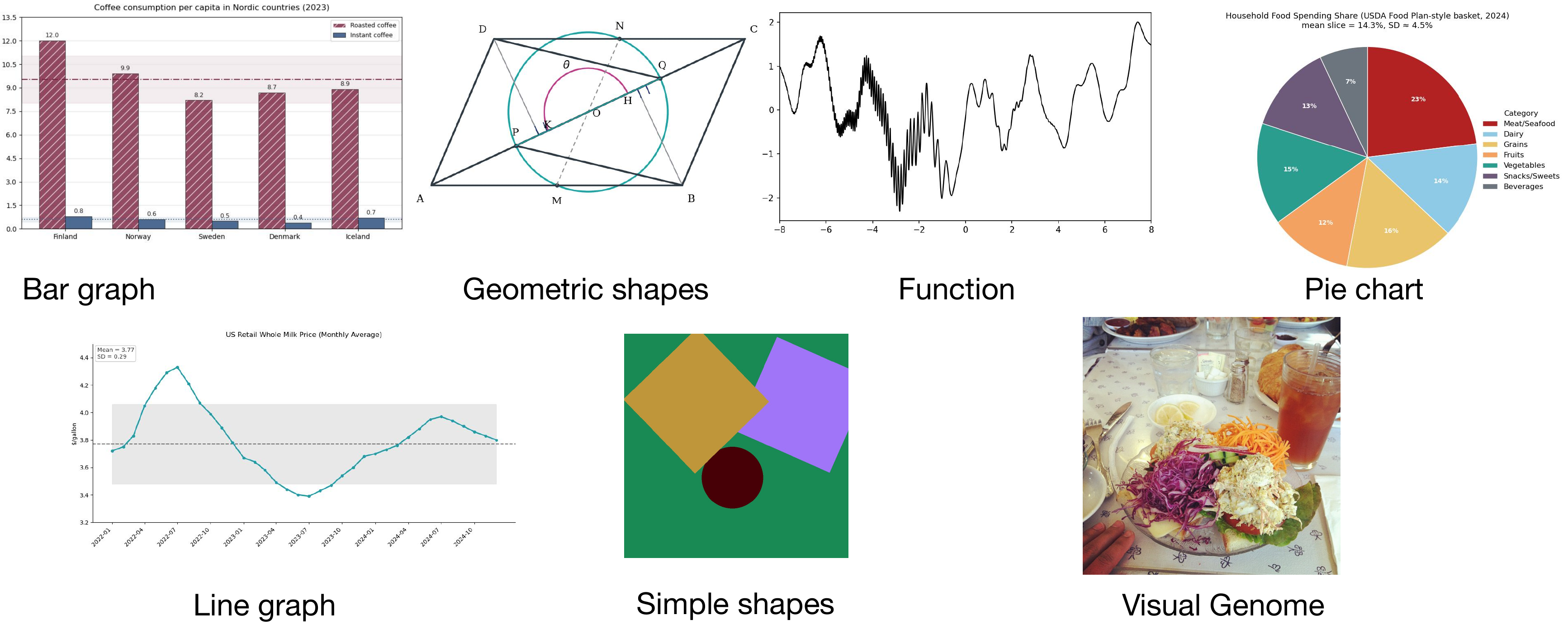}
        \vspace*{-.2cm}
    \caption{Representative target images (\textit{hard} level) for seven categories in the dataset}
    \label{fig:instances}
        \vspace*{-.3cm}
\end{figure*}

\section{Experimental Setup}
\label{sec:setup}

\subsection{Models}
\label{sec:models}

\paragraph{Describer models.}
We evaluate a set of publicly available VLMs in the Describer role.
\textsc{Qwen3-VL-30B-A3B-Instruct}~\cite{yang2025qwen3technicalreport} was run locally on a
single NVIDIA A100 GPU and \textsc{GPT-5.2} model was accessed via API. A fixed token budget of \textit{200~tokens} was used.

\paragraph{Generator models.}
We use two image generation models as generators:
\textsc{Gemini-3.1-nano-banana} and \textsc{GPT-Image-1.5}.
Both are accessed via their respective image generation APIs in
\emph{image-to-image conditioning} mode: at each turn the Generator receives
the previous rendering~$I_{t-1}$ together with the accumulated dialogue
history, not merely the latest instruction. Calls are made with default sampling parameters; seeds are not fixed across
turns (as neither API exposes turn-level seed control), so observed score
changes across turns reflect a combination of conditioning behaviour and
sampling stochasticity.
Each Describer is evaluated against each Generator, yielding a
Describer~$\times$~Generator grid of results; the Generator is held fixed
within each grid cell so that outcome differences are attributable to the
Describer's communicative quality rather than to rendering capability.

\subsection{Benchmark conditions}
\label{sec:conditions}

The benchmark comprises \textbf{14~experiment conditions} obtained by crossing
seven instance categories with two difficulty levels, \emph{easy} and
\emph{hard}.
Each condition contains \textbf{10~episodes}, yielding \textbf{140~episodes}
in total. The six categories are generated synthetically by prompting an LLM to
produce Python code, the code is executed to obtain the target image (see Figure~\ref{fig:gen_syn_prompt} in Appendix~\ref{app:prompts}).

Figure~\ref{fig:instances} shows representative target images. \textbf{Simple Shapes} contain coloured geometric primitives at specified positions; differences are visually salient but require precise spatial and color language.
\textbf{Geometry} diagrams depict labelled figures and constructions with subtle metric differences.
\textbf{Functions} show mathematical function graphs; differences concern curve shape, scale, and axis labels.
\textbf{Line Graph}, \textbf{Bar Graph}, and \textbf{Pie Graph} encode statistical data visually; corrections must be numerically precise.
We also use real photographs from the \textbf{Visual Genome}~\cite{DBLP:journals/ijcv/KrishnaZGJHKCKL17} dataset to compare the with synthetic categories. 

Difficulty is defined consistently across synthetic categories: \emph{easy}
instances use fewer elements, simpler structure, and unambiguous layouts, while
\emph{hard} instances introduce additional complexity---more data series, finer
geometric detail, denser labelling---that increases both perceptual and
verbalisation demands.  For Visual Genome, easy instances depict scenes with a small number of salient
objects; hard instances are denser and more cluttered. All instances were checked and verified manually (synthetic ones were generated with the complexity included in the prompt and then confirmed, Visual Genome instances were manually selected).
Hard instances receive consistently lower mean similarity scores than easy
instances across all categories and model pairs
(Table~\ref{tab:scores_by_category}, Appendix~\ref{app:results}), confirming
the intended difficulty ordering.

\subsection{Tasks \& Evaluation}
\label{sec:evaluation}

We evaluate episode outcomes using two tasks.

\paragraph{Task~1: Pairwise Similarity}
Judges are shown the target image (\emph{Image~A}) and a Generator
rendering (\emph{Image~B}) and asked to rate their visual and
semantic similarity on a 0--10 scale:

\begin{quote}\small
\textit{How similar is Image~B to Image~A?
Image~A is the original.
Image~B was generated by a model.
Please rate their overall visual \& semantic similarity using the scale
below:}\\[4pt]
\textbf{0} \quad Completely different image\\
\textbf{1--3} \quad Very different; only a few elements in common\\
\textbf{4--6} \quad Partial match; some key elements captured, but important differences remain\\
\textbf{7--9} \quad Strong match; most elements correct, with minor differences in detail, colour, or placement\\
\textbf{10} \quad Identical or indistinguishable
\end{quote}

\paragraph{Task~2: Triplet Preference}
Judges are shown the target image alongside two Generator renderings (from the first and the final turns)
and asked to \textit{select which rendering is more visually similar to the target}.
This two-alternative forced-choice format abstracts away from absolute scale
calibration and directly measures relative reconstruction quality---for
example, whether a later-turn rendering is preferred over an earlier one considering the iterative improvement across turns.

\paragraph{Human judges.}
We collect human ratings for both tasks by asking them to annotate all 140 episodes.
Each data point in both tasks is rated by three independent annotators with backgrounds
in Computational Linguistics and Computer Science (who are also authors).
The annotation guidelines and interfaces are given in Appendix~\ref{app:human}.

\paragraph{LLM judges.}
We use three LLMs as judges for both tasks.
\textsc{GPT-5.2} and
\textsc{Qwen3-VL-30B} are drawn from the same
model families used in the Describer role, allowing us to examine whether a
model's ratings are consistent with---or systematically biased by---its prior
exposure to the task as a generator of descriptions.
Self-serving evaluation, in which a model rates its own outputs more favourably
than an independent observer would, is a known failure mode of LLM-as-judge
pipelines~\citep{DBLP:conf/eccv/FuHLFWLRSMK24,marjieh2024large}; including
these models as judges provides a direct test of whether this bias is present
in our setting.
\textsc{Claude-4.5 Sonnet} was not used in experiments and therefore we included it as an independent, out-of-distribution LLM judge.
Comparing the three judges against each other and against human ratings allows
us to assess both the overall reliability of LLM-based evaluation in this
setting and the degree to which judge scores are confounded by a model's
participation in the task being evaluated.
All three models receive the same prompt and rating scale as human annotators.

\begin{table}[ht]
\centering
\caption{Automatic judge calibration across annotation tasks (all model pairs). \textbf{Task~1} — Pairwise Similarity: Pearson correlation ($r$) and Spearman rank correlation ($\rho$) on matched instance--turn pairs. \textbf{Task~2} — Triplet Preference: pairwise agreement rate (\%) and Cohen's $\kappa$ on binary final-vs-turn-0 choices. LipSim is treated as an additional perceptual judge.}
\label{tab:judge_calibration}
\footnotesize
\begin{tabular}{llrr}
\toprule
\multicolumn{4}{l}{\textbf{Task 1 — Pairwise Similarity}} \\
Judge A & Judge B & $r$ & $\rho$ \\
\midrule
GPT & Claude & 0.76 & 0.74 \\
GPT & Qwen3 & 0.65 & 0.65 \\
Claude & Qwen3 & 0.63 & 0.65 \\
\midrule
\multicolumn{4}{l}{\textbf{Task 2 — Triplet Preference}} \\
Judge A & Judge B & \% & $\kappa$ \\
\midrule
GPT & Claude & 62.28 & 0.24 \\
GPT & Qwen & 53.29 & 0.06 \\
Claude & Qwen & 49.10 & -0.02 \\
GPT & LipSim & 62.57 & 0.22 \\
Claude & LipSim & 52.99 & 0.05 \\
Qwen & LipSim & 57.78 & 0.15 \\
\bottomrule
\end{tabular}
\end{table}

\paragraph{Automated metrics.}
LipSim~\citep{DBLP:conf/iclr/GhazanfariAKKG24} serves as the automated metric
for Task~2; standard alternatives (SSIM, LPIPS, DINO, CLIP) are excluded as
they are well-documented to diverge from human mid-level perceptual
judgements~\citep{DBLP:conf/iclr/GhazanfariAKKG24,DBLP:conf/eccv/FuHLFWLRSMK24}.
LipSim provides a provably robust perceptual similarity distance using
Lipschitz-constrained neural networks, offering certified stability under
input perturbations.

\section{Results and Analysis}
\label{sec:results}

\subsection{Automated judge calibration}
\label{sec:results:judges}

\begin{figure}[t]
\centering
  \includegraphics[width=\columnwidth]{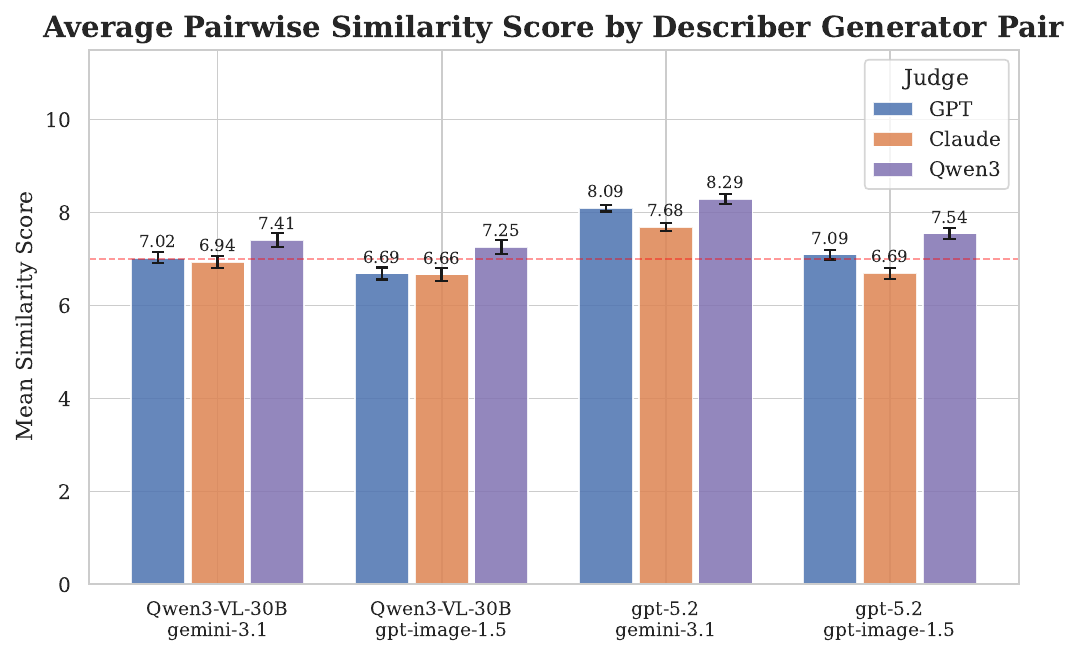}
  \caption{%
    Task-1: Mean similarity scores for all four describer--generator pairs under
    LLM judges.}
  \label{fig:main_effects}
\end{figure}

Table~\ref{tab:judge_calibration} reports pairwise judge agreement for both
evaluation tasks.
Figure~\ref{fig:main_effects} shows that all three automated
judges agree mainly on the overall ranking of model pairs, with
\textsc{gpt-5.2}~+~\textsc{gemini-3.1} consistently highest and
\textsc{Qwen3}~+~\textsc{gpt-image-1.5} consistently lowest among all judges.

\textbf{Task~1 (pairwise similarity).}
GPT-5.2 and Claude-4.5 agree more closely ($r = 0.76$, $\rho = 0.74$),
while Qwen-3 shows lower alignment with both ($r = 0.65$ / $\rho = 0.65$
vs.\ GPT-5.2; $r = 0.63$ / $\rho = 0.65$ vs.\ Claude).

\textbf{Task~2 (triplet preference).}
GPT-5.2 and Claude-4.5 reach fair agreement (62.2\%, $\kappa = 0.23$), while Qwen-3 shows lower alignment. The \textit{LipSim} aligns more closely with GPT-5.2 (62.6\%, $\kappa = 0.22$).

\subsection{Overall Performance}
\label{sec:results:findings}

Figure~\ref{fig:main_effects} shows mean similarity scores for all four
describer--generator pairs (full breakdown in
Table~\ref{tab:scores_by_category}, Appendix~\ref{app:results}).
\textbf{The describer is the dominant factor (RQ1)}: \textsc{gpt-5.2}
outperforms \textsc{Qwen-3} by approximately one point across all conditions,
with the ranking stable across all three judges.
Within each describer, \textsc{gemini-3.1} yields higher scores than
\textsc{gpt-image-1.5} (\textbf{RQ2}).
Category-level results reveal two separable bottlenecks: \emph{perception}
limits scores on \textbf{Functions} and \textbf{Geometry} hard instances
(fine-grained details such as curve shapes, axis ticks), while \emph{verbalisation} is the separate challenge of
converting perceived differences into actionable instructions.

\subsection{Iterative Refinement}
\label{sec:results:refinement}

\begin{figure}[t]
    \centering
    \includegraphics[width=1\linewidth]{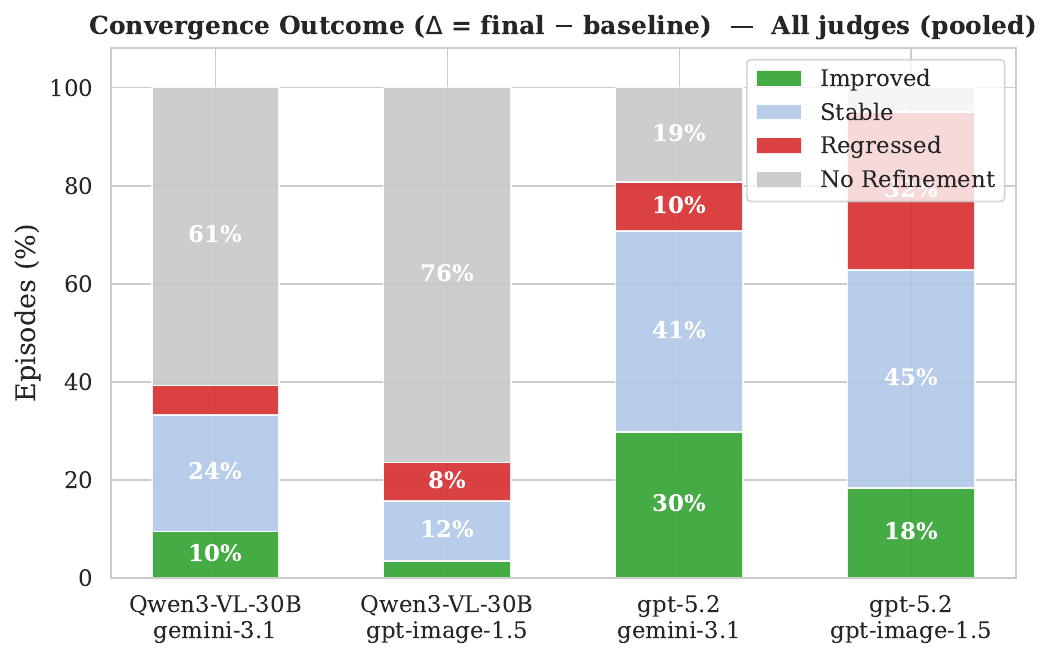}
    \caption{Convergence outcome distribution per model pair (pooled across all
    three LLM judges). \textit{Improved} ($\Delta>0$), \textit{Stable}
    ($\Delta=0$), \textit{Regressed} ($\Delta<0$), \textit{No Refinement}
    (single-turn episode).}
    \label{fig:delta_convergence}
\end{figure}

The iterative payoff (the difference in scores between the final turn image in comparison to the first turn image) is as follows and address the \textbf{RQ3}: $\Delta_Q = \text{score}_{I_T} - \text{score}_{I_1}$. \textsc{Gemini-3.1} pairings improve across turns; \textbf{\textsc{gpt-image-1.5} pairings show near-zero or negative deltas---additional turns degrade rather than improve the reconstruction} (Figures~\ref{fig:delta_final_first_turns} and~\ref{fig:triplet_final_turn_preference},
Appendix~\ref{app:results}). \textbf{The generator also determines whether iterative refinement is beneficial}. Figure~\ref{fig:delta_convergence} shows convergence outcomes per pair where each category refers to whether the first turn image has changed in scores in comparison to the final turn one. 
Qwen3-based pairs produce mostly single-turn episodes, leaving little room for improvement or regression; \textsc{gpt-5.2}/\textsc{gpt-image-1.5} continues
to the turn limit but still regresses in roughly a third of cases.
More turns do not guarantee higher quality (see also
Figure~\ref{fig:trajectory_heatmap}, Appendix~\ref{app:results}).
Among all pairs, \textsc{gpt-5.2}~+~\textsc{gemini-3.1} shows the most
consistent per-turn improvement.

\subsection{Describer Token Budget}
\label{sec:results:budget}

\begin{figure}[t]
    \centering
    \includegraphics[width=1\linewidth]{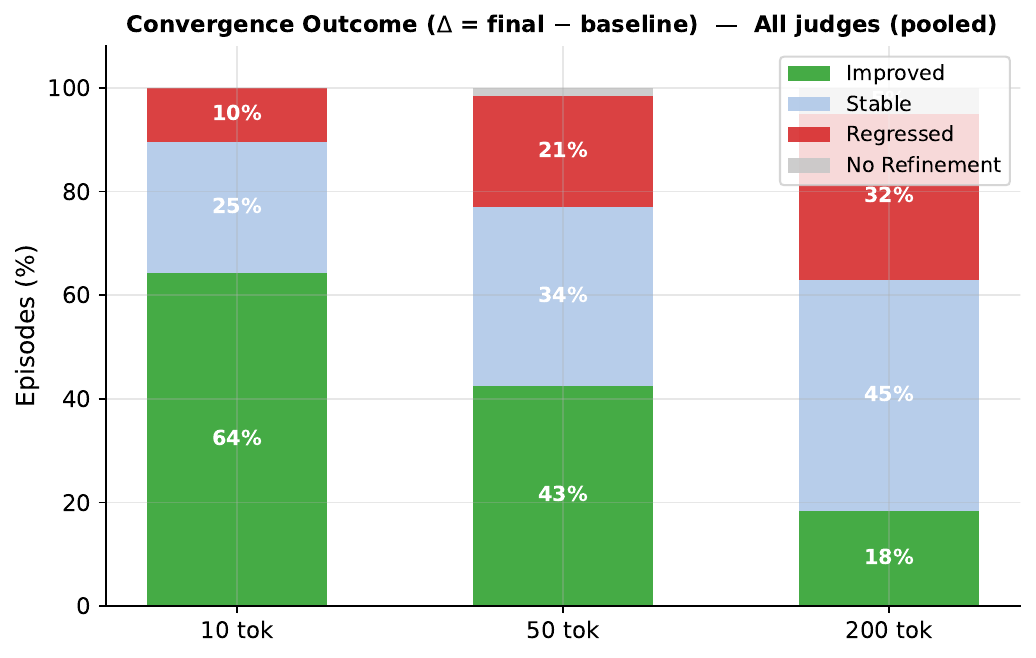}
    \caption{Convergence outcomes by token budget for
    \textsc{gpt-5.2}~+~\textsc{gpt-image-1.5}. A shorter budget produces a
    sparser first rendering, giving corrections more visible room to improve it.}
    \label{fig:token_budget_convergence}
\end{figure}

To analyse the effect of the token budget, we also ran \textsc{gpt-5.2}~+~\textsc{gpt-image-1.5} with 10, 50 tokens per turn to compare with 200 tokens experiments (see Figure~\ref{fig:token_budget_convergence}).
A shorter budget yields more visibly improved final images: at
10~tokens the first rendering~$I_1$ is sparse, so corrections have substantial room to close the gap; at 200~tokens~$I_1$ is already detailed, leaving little room for improvement---and more for regression.
Absolute final-turn scores increase with budget while triplet preference for the final image decreases because it is easier to tell apart the first rendering from the final when the budget is low (full statistics in
Figures~\ref{fig:token_scores} and~\ref{fig:token_triplet},
Appendix~\ref{app:results}).
\textbf{This quality--stability trade-off is driven by the starting-point
quality of~$I_1$}.

\subsection{Correction Language}
\label{sec:results:language}

\begin{figure}[t]
\centering
  \includegraphics[width=\columnwidth]{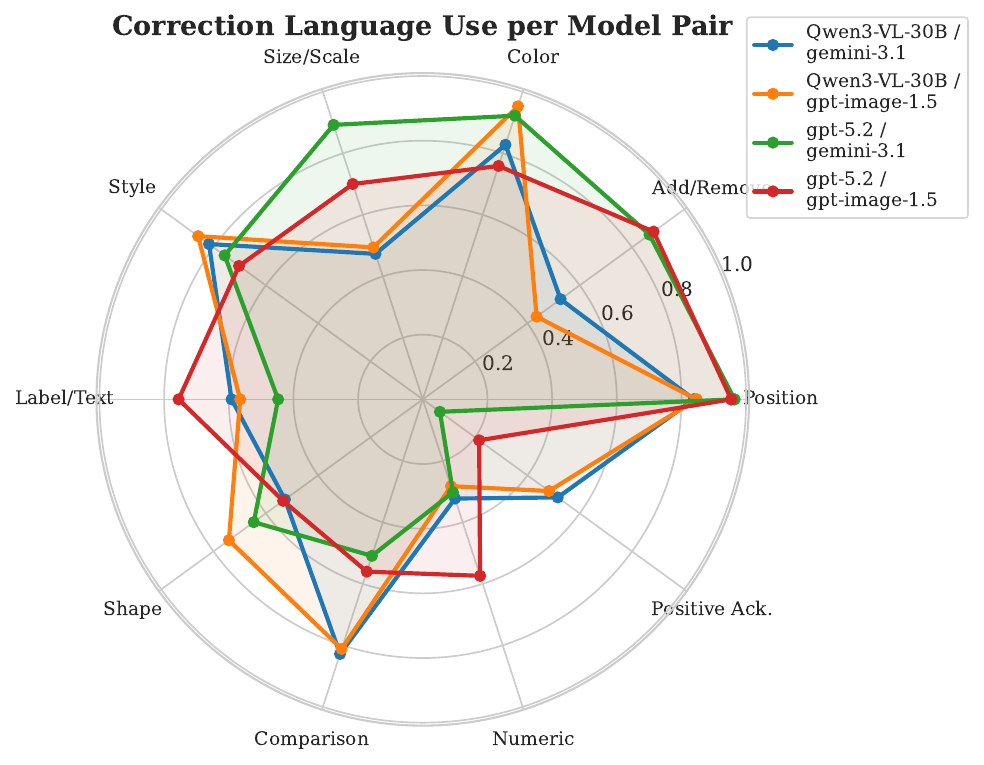}
  \caption{%
    Correction language profiles per model pair.}
  \label{fig:language_radar}
\end{figure}

We collected a list of words for respective categories that the describer uses to correct the rendering on subsequent turns. Figure~\ref{fig:language_radar} summarises the per-category usage rate per model pair. \textsc{gpt-5.2} corrections are lexically broad, spanning
\textsc{Label/Text}, \textsc{Numeric}, \textsc{Comparison}, and
\textsc{Positive Acknowledgement} in addition to the universal
\textsc{Position} and \textsc{Color} axes.
\textsc{Qwen3-VL-30B} corrections concentrate almost exclusively on
\textsc{Color} and \textsc{Shape}, with near-zero \textsc{Numeric} use
(GPT-5.2 uses coordinate notation such as \textit{x=0}, \textit{y=0};
Qwen3 contributes almost nothing) and zero \textsc{Positive Acknowledgement}. Qwen3-based pairs produce mostly single-turn episodes, so the language profile reflects only the coarsest initial instructions, with no opportunity for the structural and numeric precision that GPT-5.2 develops over longer dialogues (Table~\ref{tab:lang_keywords}: full list of keywords,  Figure~\ref{fig:top_keywords}: top-10 keywords).

\subsection{Human Judgements}
\label{sec:results:human}

Table~\ref{tab:human_calibration} reports human--automated judge calibration
for the \textsc{gpt-5.2}~+~\textsc{gpt-image-1.5} pair.
Human annotators showed moderate mutual agreement: mean $r = 0.48$ /
$\rho = 0.49$ on Task~1 and mean $\kappa = 0.29$ on Task~2
(Table~\ref{tab:human_iaa}, Appendix~\ref{app:human}).
\textbf{GPT-5.2 is the most human-aligned judge (RQ4) but still very low for practical applications}: Task-1 correlation
$r = 0.56$ is the highest among all automated judges, and Task-2 triplet
agreement is $\kappa = 0.22$. All three automated judges are systematically more tolerant than humans:
human annotators assign substantially lower mean similarity scores, with
Qwen3 showing the largest upward bias.
\textbf{Automated scores require human recalibration before they can be
used to set quality thresholds.}
Qwen3's near-chance triplet agreement with humans ($\kappa \approx 0$),
despite competitive static benchmark performance, illustrates that
game-based probing surfaces failure modes that single-turn evaluation misses.
LipSim is a practical automated complement for Task-2, matching Claude's
human agreement at lower inference cost.

\begin{table}[ht]
\centering
\caption{Human vs automatic judge calibration (\textsc{gpt-5.2/gpt-image-1.5}). \textbf{Task~1} — Pairwise Similarity: Pearson correlation ($r$) and Spearman rank correlation ($\rho$). \textbf{Task~2} — Triplet Preference: agreement rate (\%) and Cohen's $\kappa$ between Human and each LLM judge and LipSim perceptual metric.}
\label{tab:human_calibration}
\footnotesize
\begin{tabular}{llrr}
\toprule
\multicolumn{4}{l}{\textbf{Task 1 — Pairwise Similarity}} \\
Judge A & Judge B & $r$ & $\rho$ \\
\midrule
Human & GPT & 0.56 & 0.60 \\
Human & Claude & 0.51 & 0.54 \\
Human & Qwen3 & 0.41 & 0.40 \\
\midrule
\multicolumn{4}{l}{\textbf{Task 2 — Triplet Preference}} \\
Judge A & Judge B & \% & $\kappa$ \\
\midrule
Human & GPT & 60.90 & 0.22 \\
Human & Claude & 54.89 & 0.11 \\
Human & Qwen & 48.87 & -0.01 \\
Human & LipSim & 54.89 & 0.06 \\
\bottomrule
\end{tabular}
\end{table}

\subsection{Qualitative Analysis}
Qualitative samples are available in Figure~\ref{fig:qualitative1}, Figure~\ref{fig:qualitative2}, and Figure~\ref{fig:qualitative3} for \textit{Functions}, \textit{Geometry} and \textit{Visual Genome} instances. Across all three category types, \textsc{gpt-5.2}~+~\textsc{gemini-3.1} produces the closest renderings; \textsc{gpt-5.2}~+~\textsc{gpt-image-1.5} runs to ten turns but still falls short on fine details. Lower token budgets and Qwen3-based pairs yield semantically weaker reconstructions, with key structural elements missing or substituted.

\section{Conclusion}
\label{sec:conclusion}

We introduced the Image Reconstruction Game, a fully automated multimodal
reference game that turns common ground into a directly observable artefact: the rendered image produced after each dialogue turn.
Unlike static benchmarks, the game separately measures two distinct
model competencies---the ability to perceive and verbalise residual
differences (Describer), and the ability to incorporate targeted corrections without losing prior progress (Generator). Our experiments show that describer quality is the dominant factor in
reconstruction fidelity, while the generator determines whether multi-turn interaction provides any benefit at all.
Notably, stronger VLMs are capable of fine-grained visual
articulation: even for mathematically complex and geometrically unusual images,
they produce detailed, actionable correction sequences.
Functions and geometry categories nonetheless pose the greatest challenge, confirming
that fine-grained geometric and mathematical perception remains a critical
bottleneck for current VLMs---consistent with recent geometric benchmarks. Human validation reveals that automated LLM judges systematically overestimate reconstruction quality, and even the best-aligned judge remains unreliable for episode-level decisions without human recalibration.

\section*{Limitations}

The benchmark comprises 140 episodes across four Describer--Generator pairs
(10 per category--difficulty condition)---sufficient for exploratory analysis
but modest for robust category-level conclusions and cross-model
generalisation; the observed rankings may shift with additional model
pairings or larger episode counts.
Human annotations are available for a single model pair
(\textsc{gpt-5.2}~+~\textsc{gpt-image-1.5}), limiting the scope of the
human--automated correlation analysis; extending ratings to further pairs
would validate whether iterative improvement observed by LLM judges is also
perceived by human raters.
All prompts and evaluations are in English; the game's generalisation to
other languages is not assessed.
The automated judges are themselves subject to the perceptual and linguistic
biases of their training data, and their ratings should be interpreted
alongside, rather than as a replacement for, human judgements.
The benchmark currently covers seven image categories; extending it to
broader visual domains, including natural photographs at higher resolution,
remains future work.

\section*{Ethics Statement}

All images in the dataset are either synthetically generated or drawn from
Visual Genome~\citep{DBLP:journals/ijcv/KrishnaZGJHKCKL17}, a publicly
available, consent-collected dataset.
No personally identifiable information is present in the stimuli. Annotation tasks were designed to be low-risk
and non-sensitive.
The generated images produced by the game's Generator models may contain
artefacts or unintended content.
Automated evaluation with LLM judges carries the general risk that
model-specific biases propagate into benchmark rankings; we mitigate this
by using three judges from distinct model families and validating against
human ratings.
All game transcripts, model outputs, human annotation data, and evaluation
code will be publicly released to support reproducible follow-up research.


\bibliography{custom,anthology_0,anthology_1}

\appendix

\begin{figure*}[ht]

    \centering

    \includegraphics[width=\linewidth]{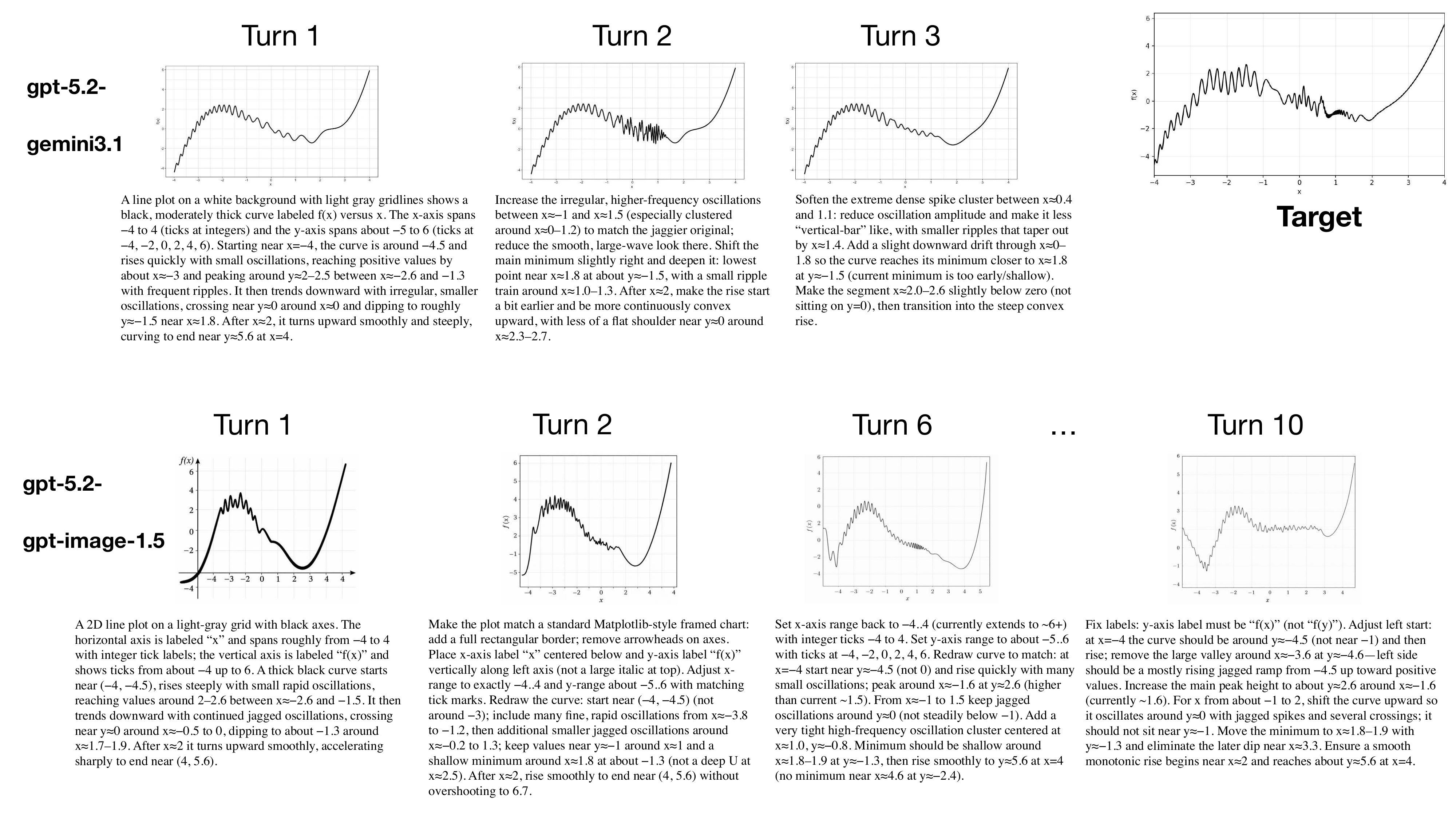}

    \vspace{0.5cm}

    \includegraphics[width=\linewidth]{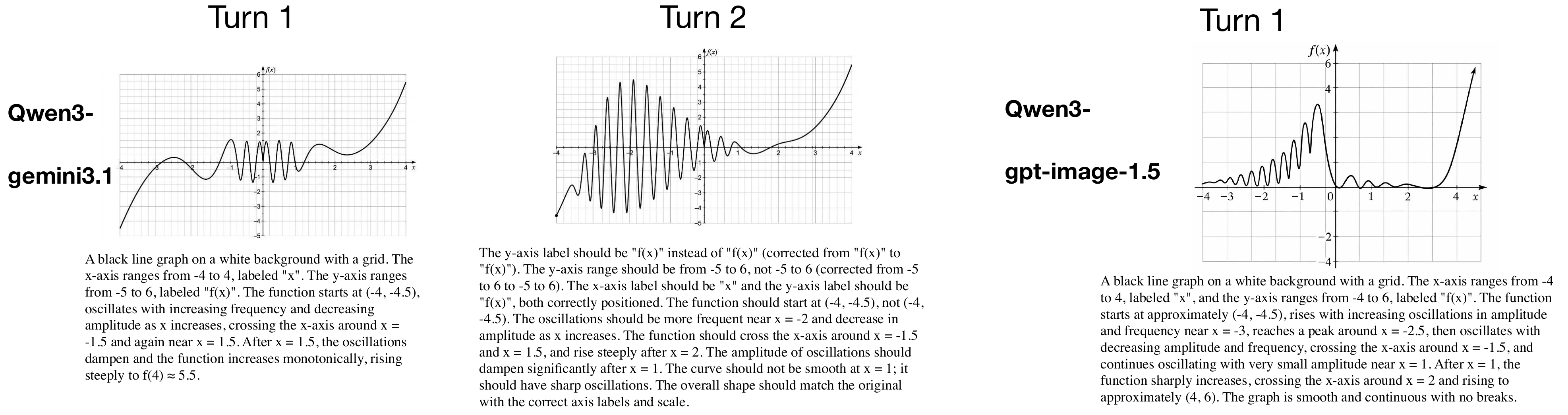}

    \vspace{0.5cm}

    \includegraphics[width=\linewidth]{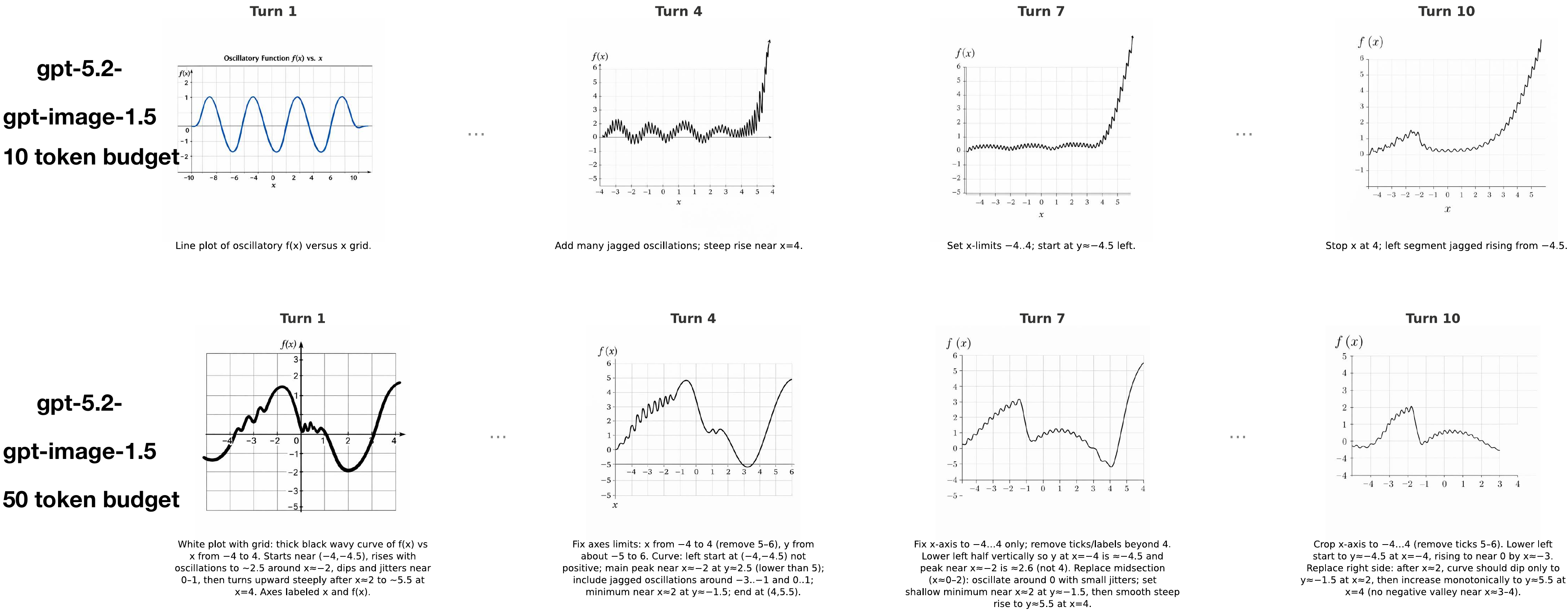}

    \caption{Qualitative samples for \textit{Functions} (hard, instance\#2) four model pairs and including token budget experiments}

    \label{fig:qualitative1}

\end{figure*}

\begin{figure*}[ht]

    \centering

    \includegraphics[width=\linewidth]{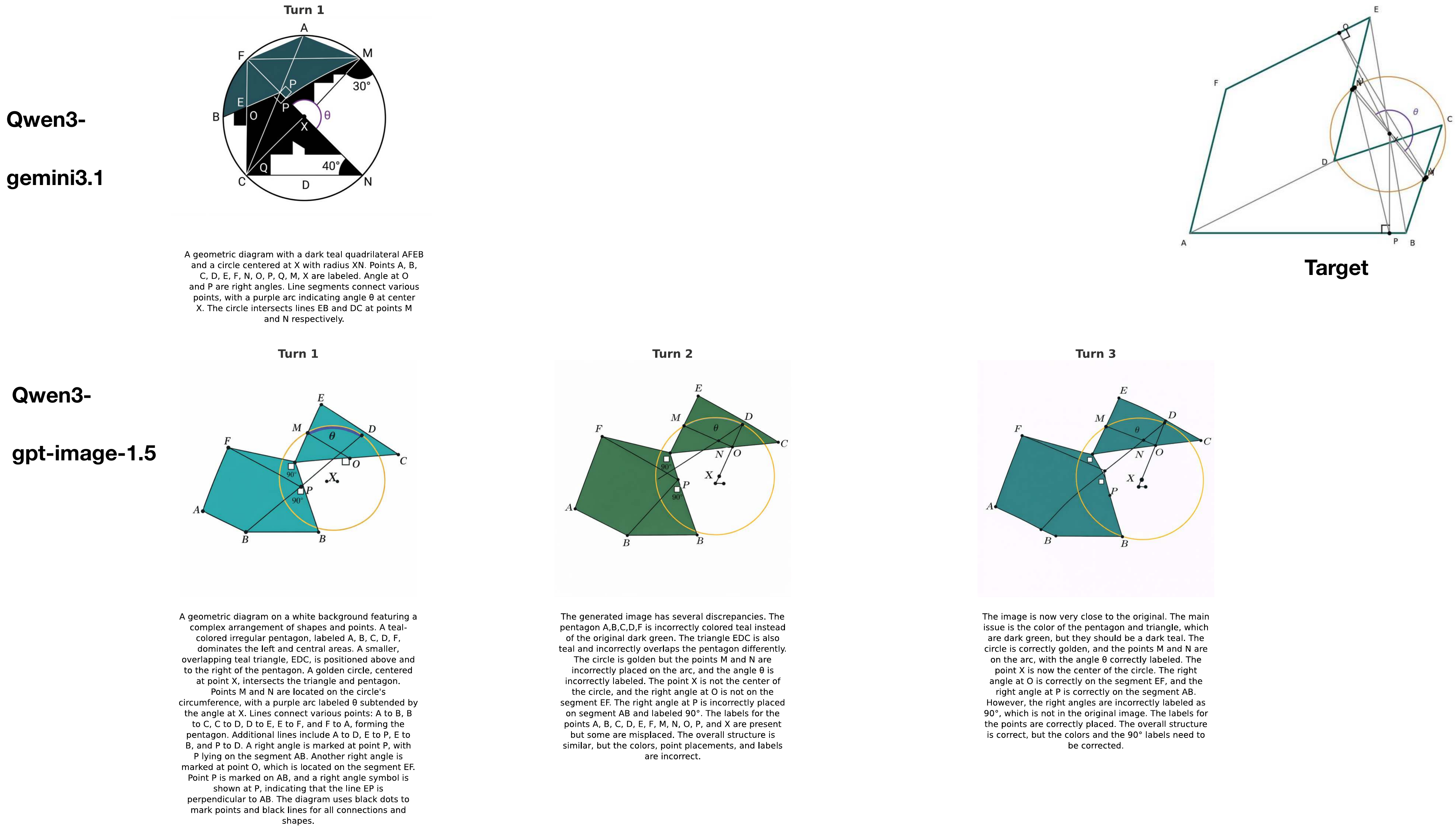}

    \vspace{0.5cm}

    \includegraphics[width=\linewidth]{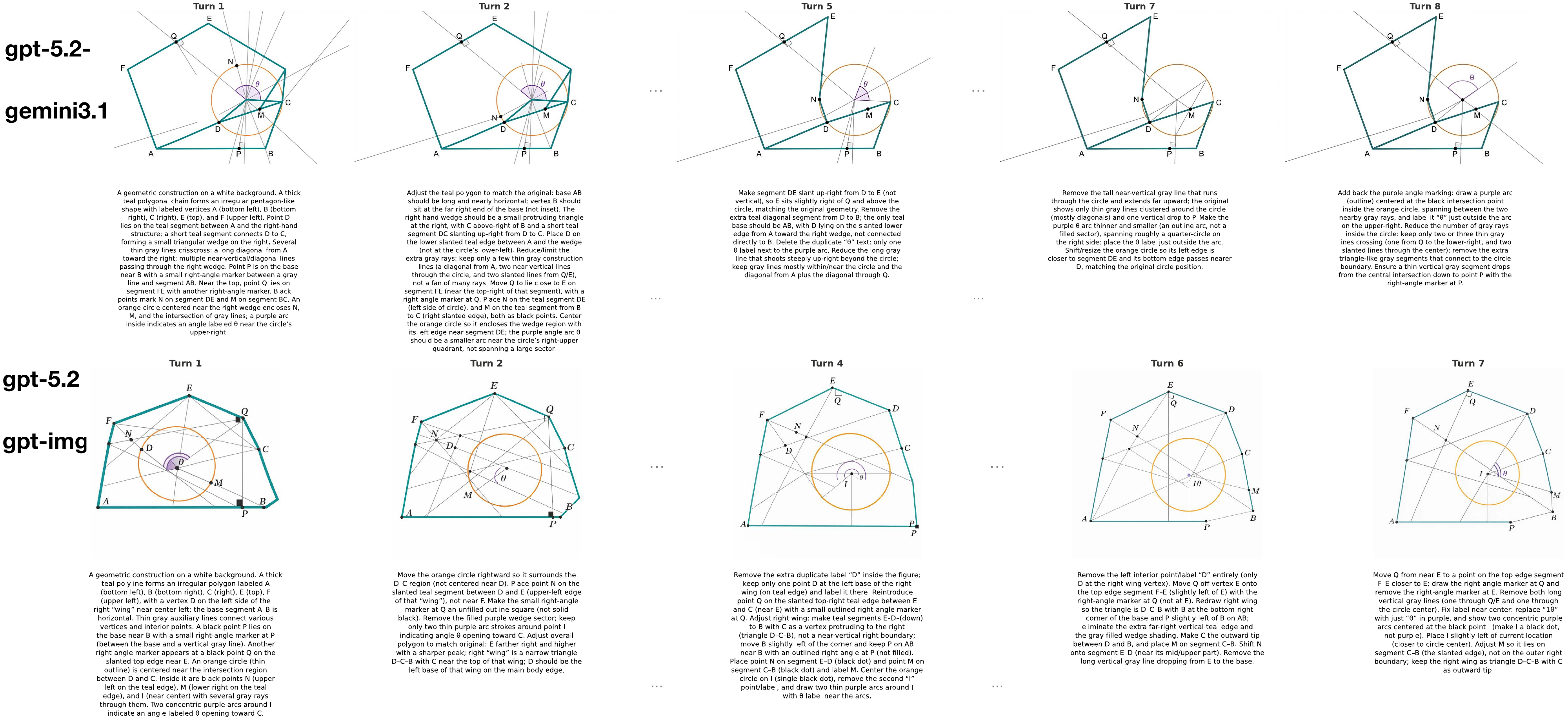}

    \vspace{0.5cm}

    \includegraphics[width=\linewidth]{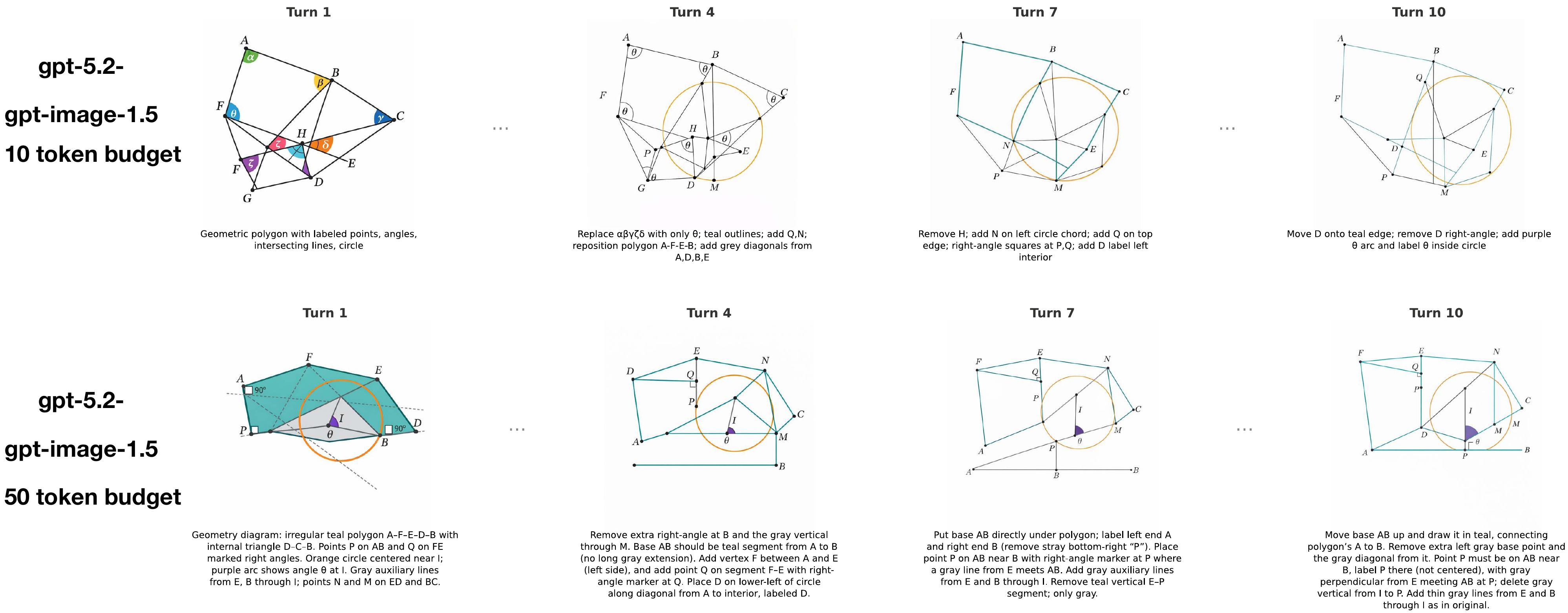}

    \caption{Qualitative samples for \textit{Geometry} (hard, instance\#3) four model pairs and including token budget experiments}

    \label{fig:qualitative2}

\end{figure*}

\begin{figure*}[ht]

    \centering

    \includegraphics[width=\linewidth]{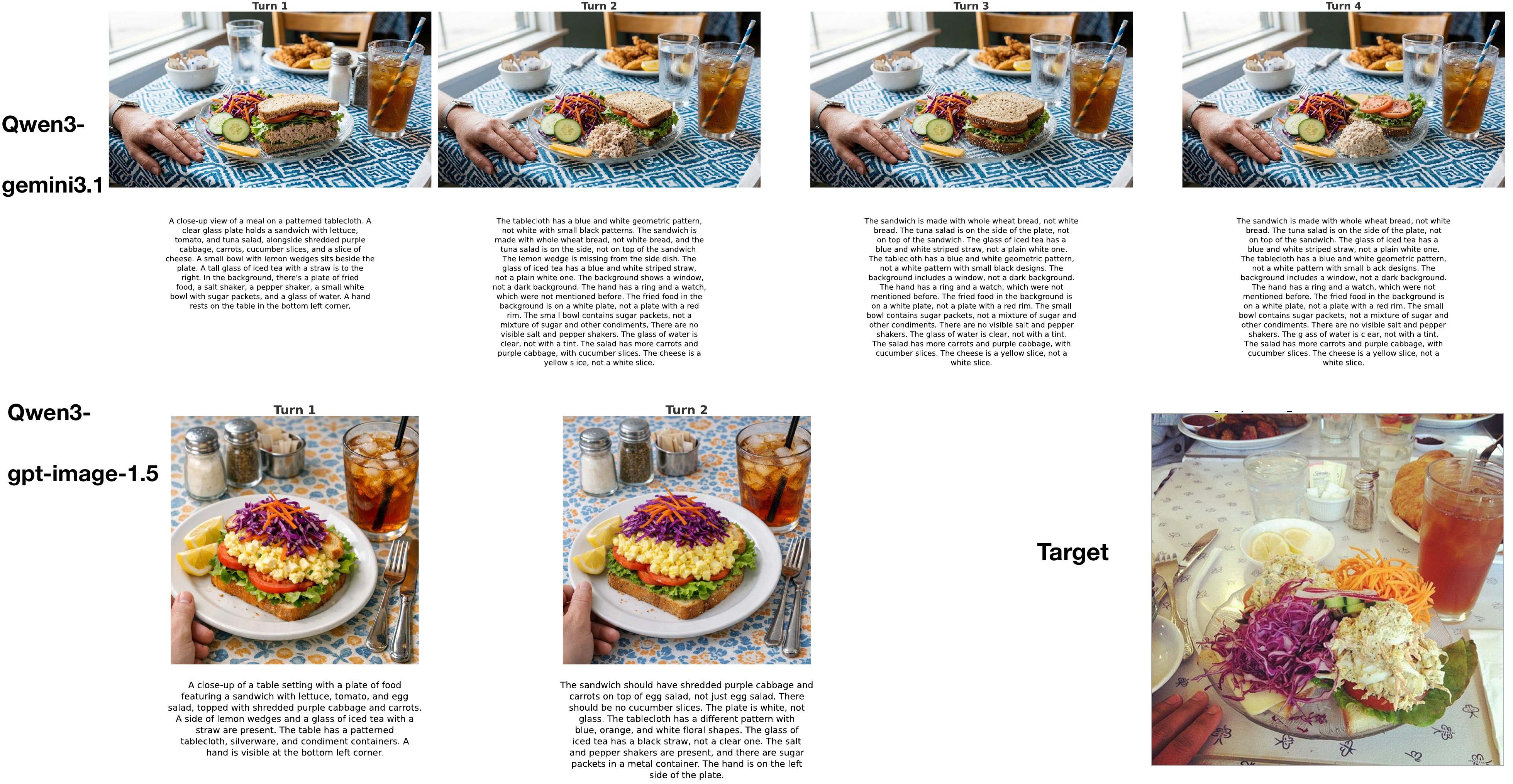}

    \vspace{0.5cm}

    \includegraphics[width=\linewidth]{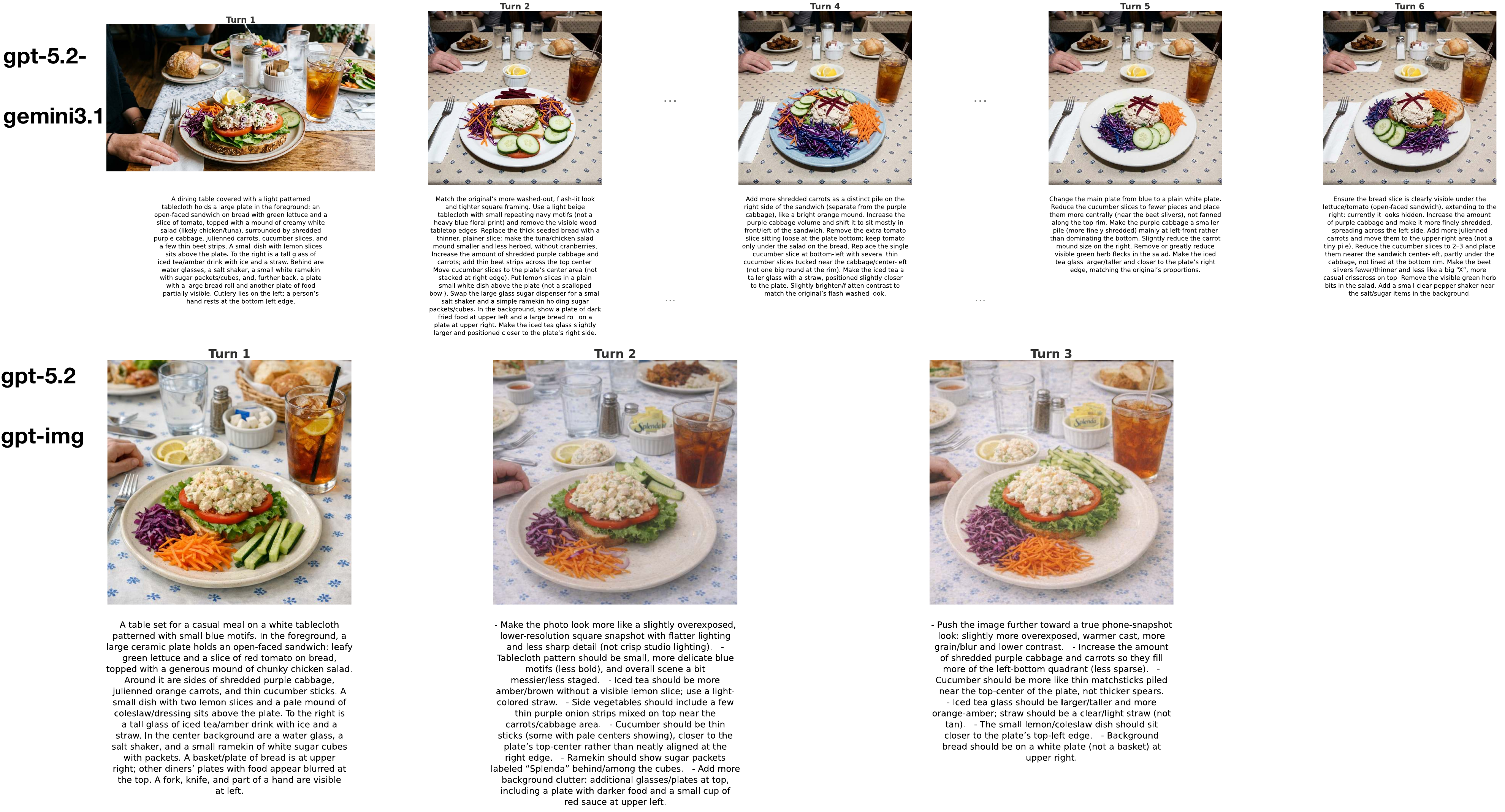}

    \vspace{0.5cm}

    \includegraphics[width=\linewidth]{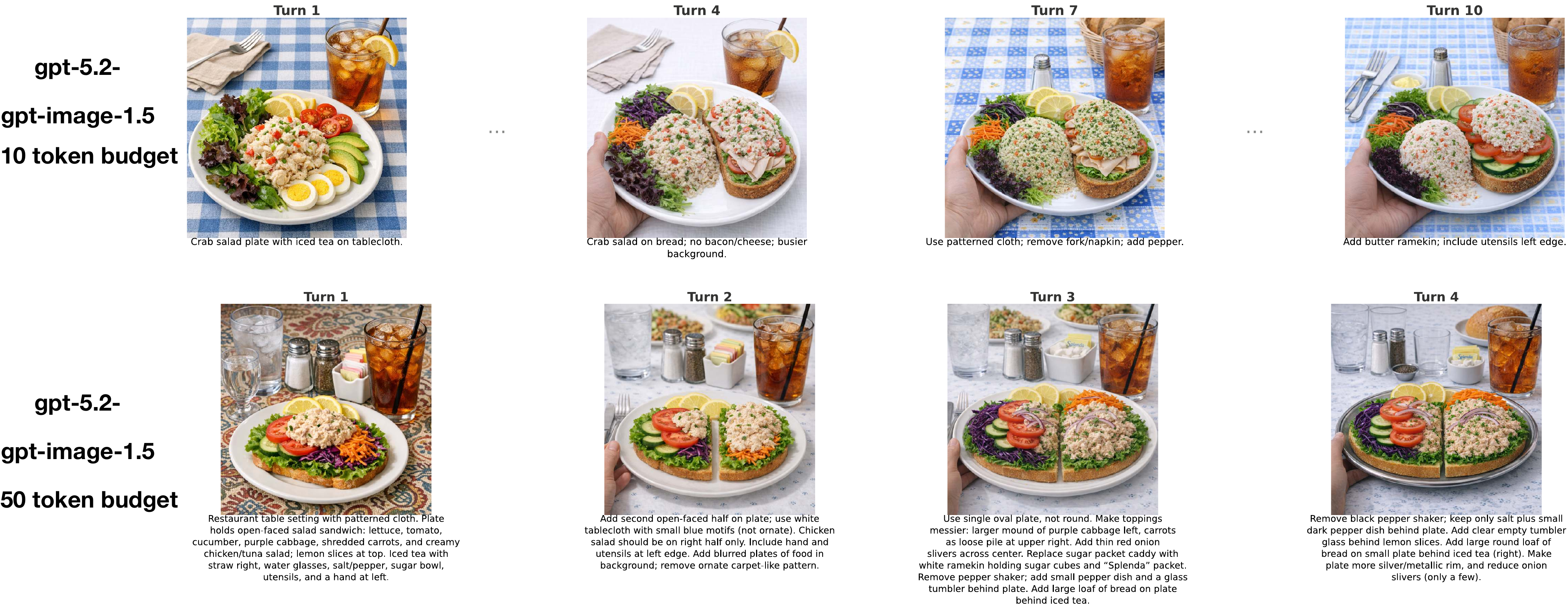}

    \caption{Qualitative samples for \textit{Visual Genome} (hard, instance\#4) four model pairs and including token budget experiments}

    \label{fig:qualitative3}

\end{figure*}

\section{Human Annotation Guidelines}
\label{app:human}

Human annotations were collected for both evaluation tasks (Section~\ref{sec:evaluation})
on the \textsc{gpt-5.2}~+~\textsc{gpt-image-1.5} model pair (140 episodes).
Three annotators with backgrounds in Computer
Science rated each episode independently; all 140 episodes were covered by
all three annotators.

\subsection*{Task~1 — Pairwise Similarity Rating}

\begin{figure*}[ht]
  \centering
  \includegraphics[width=0.95\linewidth]{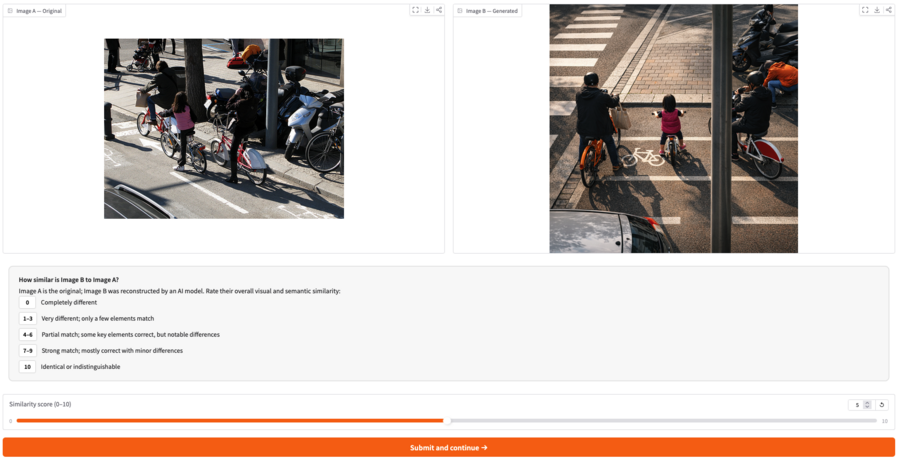}
  \caption{%
    Task~1 interface: target (\emph{Image~A}) and Generator rendering
    (\emph{Image~B}) side by side; anchor labels calibrate the 0--10 slider.}
  \label{fig:anno_ui_task1}

  \vspace{2em}

  \includegraphics[width=0.95\linewidth]{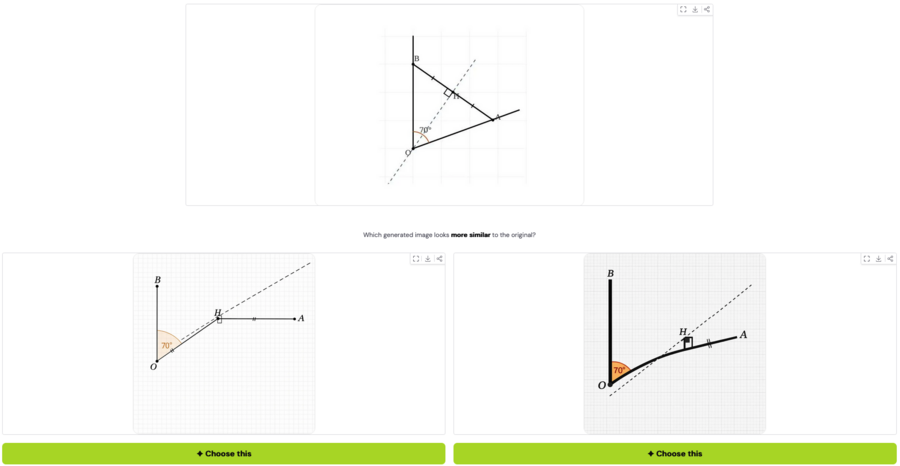}
  \caption{%
    Task~2 interface: annotators choose which of two renderings
    (first vs.\ final turn) is more similar to the target.}
  \label{fig:anno_ui_task2}
\end{figure*}

Annotators saw the target image~$I^*$ alongside each Generator rendering and
rated visual and semantic similarity on a 0--10 integer scale.
Figure~\ref{fig:anno_ui_task1} shows the interface.

\subsection*{Task~2 — Triplet Preference}

Annotators saw the target image alongside the turn-0 and final-turn
renderings and selected which was more similar to the target.
Figure~\ref{fig:anno_ui_task2} shows the interface.

\section{Additional Results}
\label{app:results}

\subsection*{Per-category similarity scores}

Table~\ref{tab:scores_by_category} reports mean similarity scores for every category
at both difficulty levels across all four judges.

\begin{table*}[ht]
\centering
\caption{Mean final-turn similarity score (1--10) per model pair, image category, and difficulty level, broken down by judge. LipSim reports \% of instances where the final-turn image was preferred over turn-0 (triplet task). Human annotations are available for the gpt-5.2\,/\,gpt-image-1.5 pair only; `--' indicates no annotation.}
\label{tab:scores_by_category}
\footnotesize
\setlength{\tabcolsep}{4pt}
\begin{tabular}{l l l r r r r r}
\toprule
\textbf{Model Pair} & \textbf{Category} & \textbf{Diff.} & \textbf{GPT} & \textbf{Claude} & \textbf{Qwen3} & \textbf{LipSim (\%)} & \textbf{Human} \\
\midrule
\multirow{14}{*}{\textbf{Qwen-3 / gemini-3.1}} & Bar Graph & easy & 8.70 & 8.70 & 9.00 & --- & {--} \\
 & Bar Graph & hard & 8.50 & 8.10 & 8.80 & 0.0\% & {--} \\
 & Functions & easy & 6.10 & 6.70 & 7.40 & 100.0\% & {--} \\
 & Functions & hard & 5.20 & 5.60 & 6.00 & 40.0\% & {--} \\
 & Geometry & easy & 7.20 & 6.40 & 6.50 & 40.0\% & {--} \\
 & Geometry & hard & 5.20 & 5.00 & 5.10 & 33.3\% & {--} \\
 & Line Graph & easy & 8.00 & 7.90 & 8.50 & 50.0\% & {--} \\
 & Line Graph & hard & 6.90 & 6.90 & 7.10 & 0.0\% & {--} \\
 & Pie Graph & easy & 7.90 & 7.90 & 8.90 & 66.7\% & {--} \\
 & Pie Graph & hard & 8.10 & 8.30 & 8.40 & 33.3\% & {--} \\
 & Simple Shapes & easy & 6.50 & 5.90 & 6.70 & 66.7\% & {--} \\
 & Simple Shapes & hard & 7.00 & 6.30 & 7.50 & 33.3\% & {--} \\
 & Visual Genome & easy & 6.40 & 6.70 & 7.30 & 55.6\% & {--} \\
 & Visual Genome & hard & 6.60 & 6.70 & 6.50 & 44.4\% & {--} \\
\midrule \\
\multirow{14}{*}{\textbf{Qwen-3 / gpt-image-1.5}} & Bar Graph & easy & 7.70 & 7.60 & 8.80 & --- & {--} \\
 & Bar Graph & hard & 7.60 & 7.10 & 8.20 & --- & {--} \\
 & Functions & easy & 4.90 & 4.40 & 6.20 & 100.0\% & {--} \\
 & Functions & hard & 4.80 & 5.90 & 7.40 & 100.0\% & {--} \\
 & Geometry & easy & 6.30 & 5.90 & 5.70 & 0.0\% & {--} \\
 & Geometry & hard & 5.20 & 5.10 & 5.30 & 100.0\% & {--} \\
 & Line Graph & easy & 7.10 & 7.50 & 8.00 & --- & {--} \\
 & Line Graph & hard & 6.70 & 7.20 & 7.20 & 100.0\% & {--} \\
 & Pie Graph & easy & 8.20 & 8.10 & 8.40 & 0.0\% & {--} \\
 & Pie Graph & hard & 8.20 & 8.00 & 7.90 & 0.0\% & {--} \\
 & Simple Shapes & easy & 6.80 & 6.60 & 7.10 & 40.0\% & {--} \\
 & Simple Shapes & hard & 6.60 & 6.60 & 6.50 & 16.7\% & {--} \\
 & Visual Genome & easy & 7.20 & 7.00 & 8.10 & 0.0\% & {--} \\
 & Visual Genome & hard & 6.30 & 6.30 & 6.70 & 40.0\% & {--} \\
\midrule \\
\multirow{14}{*}{\textbf{gpt-5.2 / gemini-3.1}} & Bar Graph & easy & 8.90 & 8.80 & 9.20 & 66.7\% & {--} \\
 & Bar Graph & hard & 8.80 & 8.60 & 9.20 & 100.0\% & {--} \\
 & Functions & easy & 7.30 & 7.00 & 7.20 & 88.9\% & {--} \\
 & Functions & hard & 6.50 & 7.20 & 6.70 & 66.7\% & {--} \\
 & Geometry & easy & 8.70 & 7.70 & 8.80 & 87.5\% & {--} \\
 & Geometry & hard & 7.90 & 6.50 & 7.50 & 88.9\% & {--} \\
 & Line Graph & easy & 8.50 & 8.00 & 9.10 & 83.3\% & {--} \\
 & Line Graph & hard & 8.00 & 8.10 & 8.60 & 88.9\% & {--} \\
 & Pie Graph & easy & 8.80 & 8.70 & 9.20 & 87.5\% & {--} \\
 & Pie Graph & hard & 8.80 & 8.60 & 8.80 & 90.0\% & {--} \\
 & Simple Shapes & easy & 7.60 & 6.60 & 8.00 & 50.0\% & {--} \\
 & Simple Shapes & hard & 7.70 & 7.00 & 7.80 & 40.0\% & {--} \\
 & Visual Genome & easy & 7.80 & 7.40 & 8.00 & 40.0\% & {--} \\
 & Visual Genome & hard & 8.00 & 7.30 & 8.00 & 80.0\% & {--} \\
\midrule \\
\multirow{14}{*}{\textbf{gpt-5.2 / gpt-image-1.5}} & Bar Graph & easy & 7.90 & 7.40 & 8.60 & 40.0\% & 6.00 \\
 & Bar Graph & hard & 7.40 & 7.00 & 8.50 & 40.0\% & 5.17 \\
 & Functions & easy & 6.10 & 5.90 & 7.20 & 80.0\% & 5.10 \\
 & Functions & hard & 4.80 & 5.20 & 6.80 & 70.0\% & 5.33 \\
 & Geometry & easy & 7.50 & 6.80 & 7.10 & 80.0\% & 5.63 \\
 & Geometry & hard & 6.60 & 5.70 & 6.20 & 100.0\% & 4.37 \\
 & Line Graph & easy & 6.60 & 6.70 & 8.20 & 80.0\% & 6.37 \\
 & Line Graph & hard & 6.00 & 6.40 & 6.90 & 80.0\% & 5.40 \\
 & Pie Graph & easy & 8.30 & 8.00 & 8.40 & 70.0\% & 6.93 \\
 & Pie Graph & hard & 7.60 & 7.30 & 7.80 & 30.0\% & 5.97 \\
 & Simple Shapes & easy & 7.50 & 6.30 & 7.10 & 42.9\% & 6.40 \\
 & Simple Shapes & hard & 7.50 & 6.80 & 7.20 & 37.5\% & 7.17 \\
 & Visual Genome & easy & 7.90 & 7.10 & 7.90 & 40.0\% & 7.47 \\
 & Visual Genome & hard & 7.60 & 7.10 & 7.70 & 55.6\% & 6.83 \\
\bottomrule
\end{tabular}
\end{table*}

\subsection*{Iterative payoff and triplet preference}

Figure~\ref{fig:delta_final_first_turns} shows $\Delta_Q$ per model pair and
judge: \textsc{gemini-3.1} pairings yield consistent positive gains while
\textsc{gpt-image-1.5} pairings show near-zero or negative deltas.
Figure~\ref{fig:triplet_final_turn_preference} corroborates this via Task-2
preference rates, with \textsc{gemini-3.1} pairings consistently above the
50\% chance baseline and \textsc{gpt-image-1.5} pairings at or below it.


\subsection*{Correction language analysis}

Correction utterances (turns $\geq 2$) were labelled across ten categories
using stem patterns (\texttt{\textbackslash b<stem>\textbackslash w\{0,6\}\textbackslash b})
and phrase patterns for multi-word expressions; a category is flagged if any
pattern fires.
Tables~\ref{tab:human_iaa} and~\ref{tab:lang_keywords} appear together on
the following page.

\begin{table*}[p]
\centering

\begin{minipage}{\textwidth}
\centering
\caption{Human inter-annotator agreement across both tasks (140 episodes,
  \textsc{gpt-5.2}~/~\textsc{gpt-image-1.5} pair).
  Task~1: Pearson $r$ and Spearman $\rho$ on 0--10 similarity scores.
  Task~2: agreement rate and Cohen's $\kappa$ on binary triplet choices.}
\label{tab:human_iaa}
\footnotesize
\begin{tabular}{lrrrr}
\toprule
 & \multicolumn{2}{c}{\textbf{Task 1}} & \multicolumn{2}{c}{\textbf{Task 2}} \\
\cmidrule(lr){2-3}\cmidrule(lr){4-5}
Pair & $r$ & $\rho$ & Agree.\ (\%) & $\kappa$ \\
\midrule
A vs.~B & 0.421 & 0.447 & 63.6 & 0.247 \\
A vs.~C & 0.562 & 0.546 & 65.7 & 0.308 \\
B vs.~C & 0.454 & 0.465 & 66.4 & 0.316 \\
\midrule
\textit{Mean} & \textit{0.479} & \textit{0.486} & \textit{65.2} & \textit{0.290} \\
\bottomrule
\end{tabular}
\end{minipage}

\vspace{1.2em}

\begin{minipage}{\textwidth}
\centering
\caption{Correction language categories with representative stem roots and
  phrase patterns. Stems are matched as
  \texttt{\textbackslash b<stem>\textbackslash w\{0,6\}\textbackslash b}
  to cover inflected forms; a category is flagged if any pattern fires.}
\label{tab:lang_keywords}
\scriptsize
\setlength{\tabcolsep}{4pt}
\begin{tabular}{lp{7.0cm}p{6.8cm}}
\toprule
Category & Representative stem roots & Representative phrase patterns \\
\midrule
Position
  & \textit{mov, shift, left, right, top, bottom, center, centr, above, below, higher, lower, align, overlap, corner, diagonal, horizon, vertic, perpendicular, parallel}
  & ``move to the left/right/top/bottom''; ``upper-left''; ``top-right corner''; ``to the left of'' \\
Size/Scale
  & \textit{enlarg, larg, bigg, small, shrink, reduc, widen, narrow, thin, thick, tall, shorter, resize, rescal, upscal, downscal}
  & ``too big/small/wide/narrow''; ``scale up/down''; ``make it larger/smaller''; ``slightly narrower'' \\
Color
  & named colours (\textit{red, blue, green, orange, purple, yellow, cyan, teal, gray, black, white, \ldots}); \textit{darker, lighter, brighter, saturate, desaturat, muted, vivid, pale}
  & ``color/colour''; ``shade of''; ``hue''; ``more saturated/vivid''; ``make it darker/lighter'' \\
Add/Remove
  & \textit{add, remov, delet, insert, includ, exclud, omit, missing, extra, eliminat, replac, substitut, absent, introduc, eras, drop}
  & ``should not be there''; ``add/insert a''; ``remove/delete the''; ``no longer visible'' \\
Shape
  & \textit{round, circular, ellips, oval, squar, rectangl, triangl, polygon, diamond, arc, curv, straight, angl, vertex, wedge, convex, concave, symmet, wavy}
  & ``should be a circle/square/triangle''; ``more rounded/pointed''; ``right-angle'' \\
Label/Text
  & \textit{label, mislabel, title, annot, font, legend, caption, tick, axis, note, header, bold, italic, underlin, misalign, misplac}
  & ``x/y-axis''; ``axis label/tick/title''; ``missing/wrong label''; ``font size/weight/style'' \\
Style
  & \textit{gradient, shadow, blur, smooth, texture, flat, solid, dashed, dotted, stroke, opaci, transparent, fill, outlin, border, grid, background, blend}
  & ``flat color/fill''; ``solid/dashed/dotted line''; ``line weight/thickness''; ``stroke width'' \\
Comparison
  & \textit{incorrect, inaccurat, erroneous, mismatch, discrepancy, differ, wrong, misrepresent}
  & ``instead of''; ``rather than''; ``should be''; ``doesn't match''; ``compared to the original'' \\
Numeric
  & \textit{approximat, exactl, precisel, percent, proportion}
  & digits with units (\%, px, pt, degrees); coordinate notation (\texttt{x=3.2}, \texttt{y$\approx$1}); ``about/around/exactly \textit{N}'' \\
Positive Ack.
  & \textit{correct, accurat, almost, nearly, well}
  & ``close/similar enough''; ``otherwise correct/good''; ``other than that''; ``now correct'' \\
\bottomrule
\end{tabular}
\end{minipage}

\end{table*}

\begin{figure*}[p]
\centering
  \includegraphics[width=\textwidth]{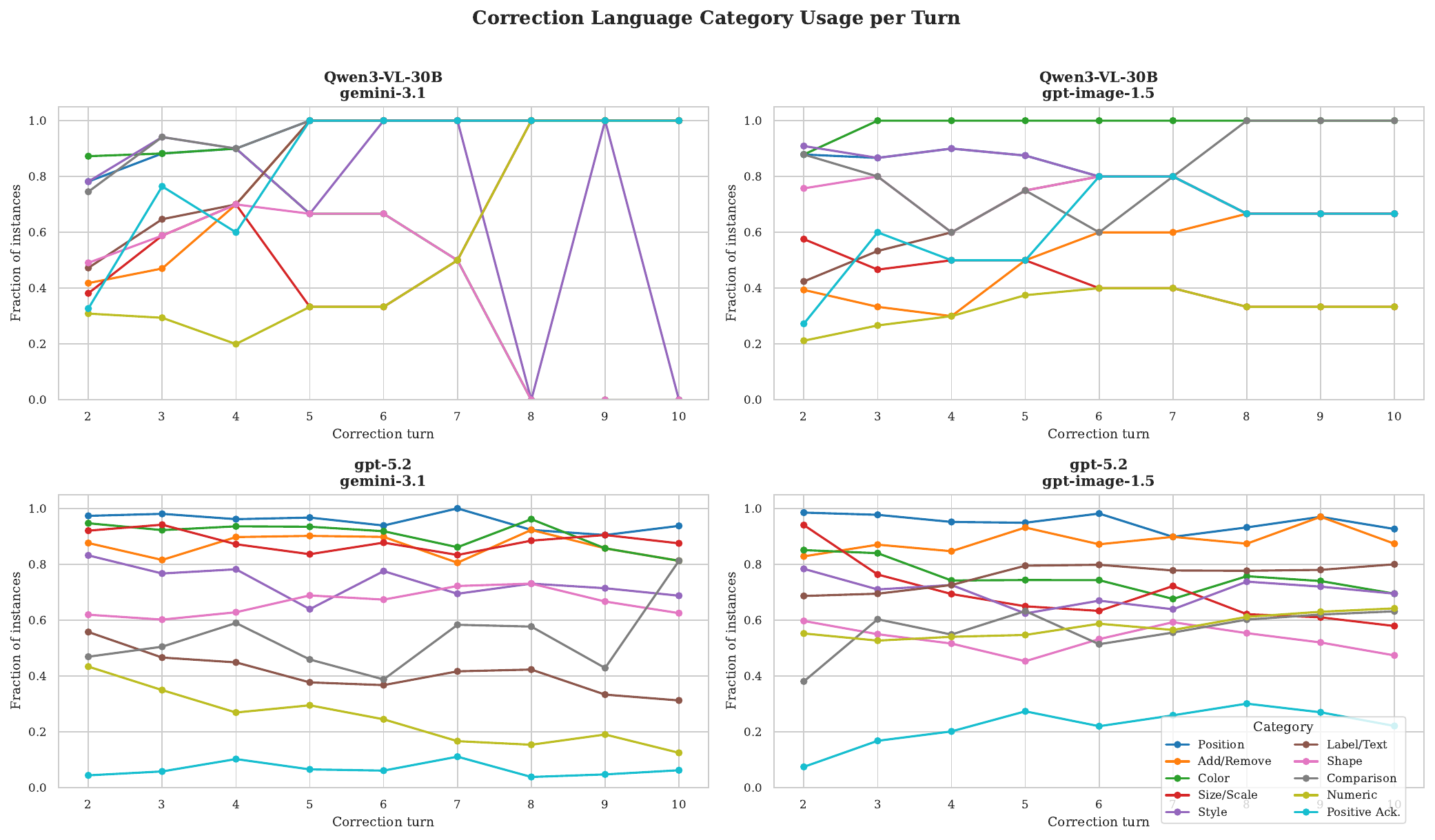}
  \caption{%
    Correction language category usage per turn for each model pair.
    \textsc{gpt-5.2} pairs (bottom row) sustain consistent multi-category
    coverage; \textsc{Qwen3-VL-30B} pairs (top row) show higher variance
    and sparse later-turn coverage.}
  \label{fig:language_turns}
\end{figure*}

\begin{figure*}[p]
\centering
  \includegraphics[width=\textwidth]{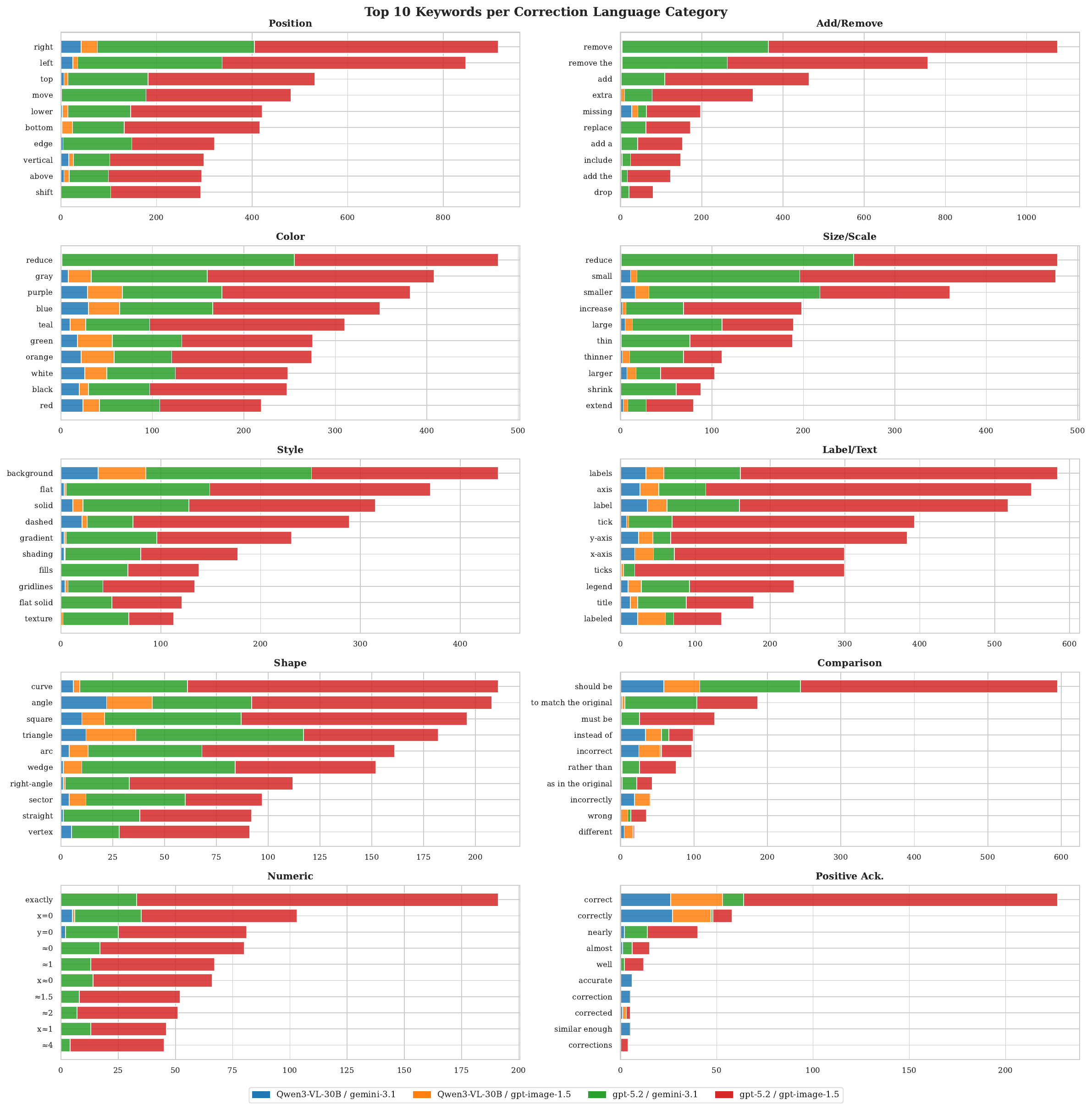}
  \caption{%
    Top-10 matched keyword tokens per correction language category by model
    pair. GPT-5.2 pairings dominate \textsc{Numeric}, \textsc{Label/Text},
    and \textsc{Positive Ack.}; Qwen3-VL-30B pairings lead in \textsc{Color}
    and \textsc{Shape} and are absent from \textsc{Numeric} and
    \textsc{Positive Ack.}}
  \label{fig:top_keywords}
\end{figure*}

Figure~\ref{fig:language_turns} shows per-category usage rates across turns.
\textsc{gpt-5.2} pairs sustain broad-coverage profiles throughout all ten
turns; \textsc{Qwen3-VL-30B} pairs show narrowing toward \textsc{Color}
dominance at later turns, partly reflecting the small number of episodes that
reach them. Figure~\ref{fig:top_keywords} shows the top-10 keyword tokens per
category: the sharpest contrasts are in \textsc{Numeric} (coordinate notation
absent in Qwen3), \textsc{Label/Text} (structural references absent in Qwen3),
and \textsc{Positive Acknowledgement} (zero counts for Qwen3).

\subsection*{Effect of describer token budget}

\begin{figure*}[p]
\centering
\begin{minipage}[t]{0.48\textwidth}
  \centering
  \includegraphics[width=\linewidth]{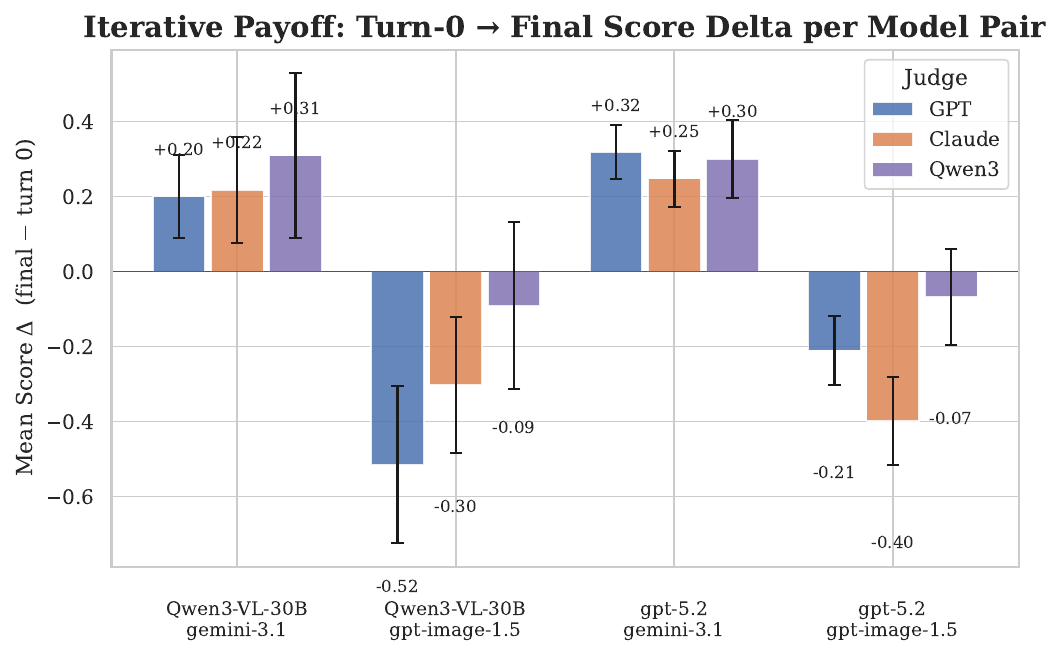}
  \caption{%
    Task-1: Mean iterative payoff $\Delta_Q = \text{score}_{I_T} - \text{score}_{I_1}$
    per model pair and judge (error bars: SEM).
    The dashed line marks the no-improvement baseline.}
  \label{fig:delta_final_first_turns}
\end{minipage}\hfill
\begin{minipage}[t]{0.48\textwidth}
  \centering
  \includegraphics[width=\linewidth]{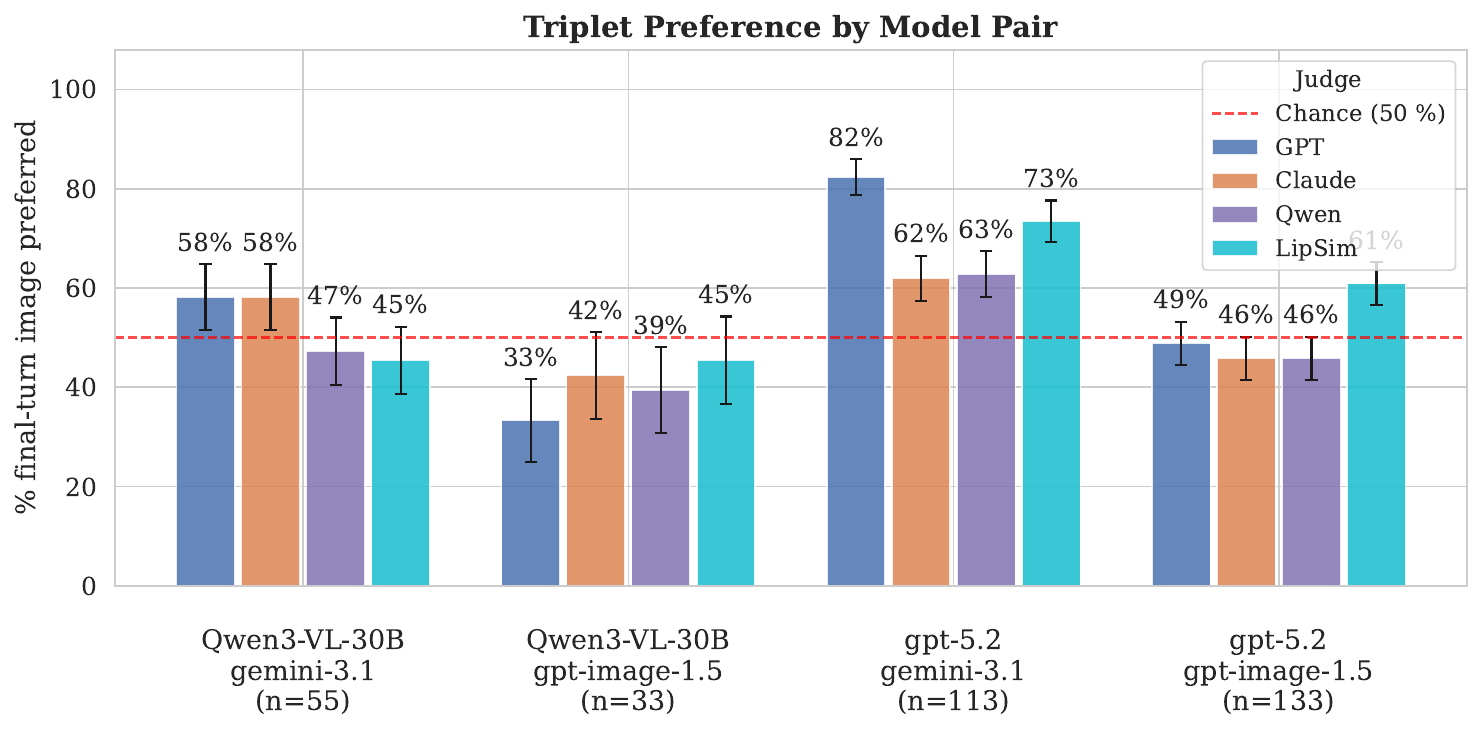}
  \caption{%
    Task-2: Percentage of multi-turn episodes where the final-turn image is
    preferred over the first-turn image, by model pair and judge.
    The red dashed line marks the 50\% chance baseline.}
  \label{fig:triplet_final_turn_preference}
\end{minipage}

\bigskip\bigskip

\begin{minipage}[t]{0.48\textwidth}
  \centering
  \includegraphics[width=\linewidth]{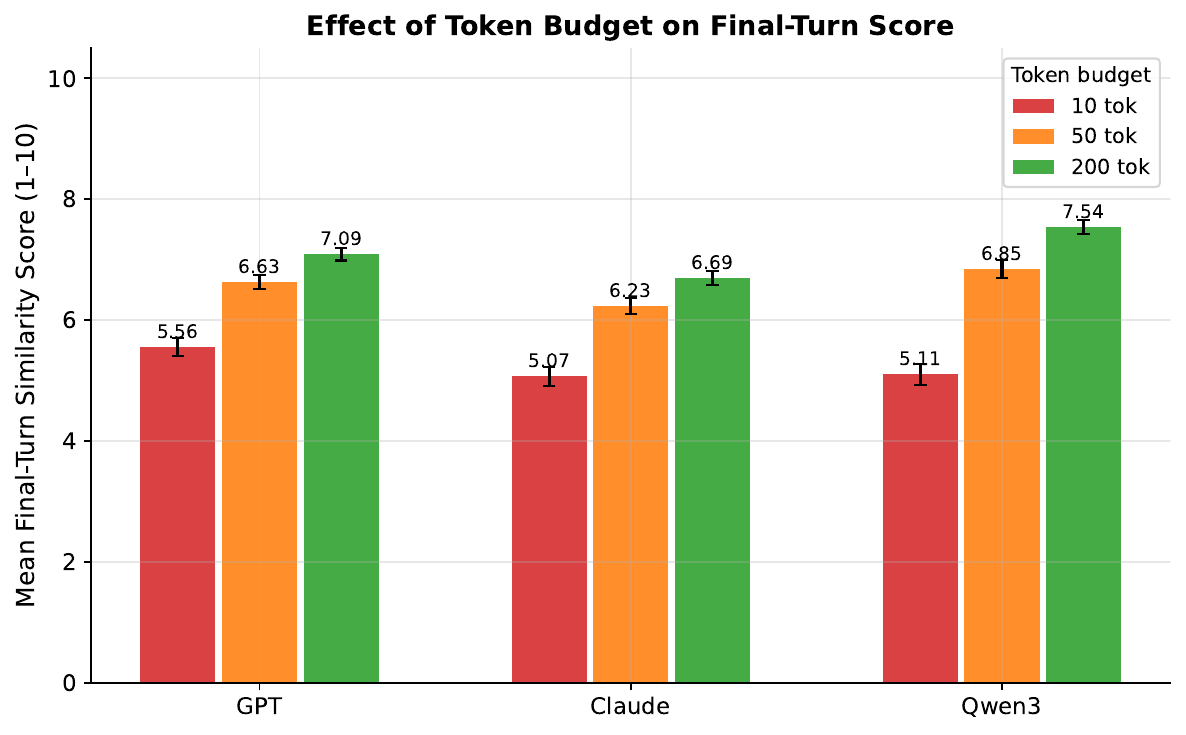}
  \caption{%
    Mean final-turn similarity score for \textsc{gpt-5.2}~+~\textsc{gpt-image-1.5}
    at three token budgets, by judge (error bars: SEM).
    Absolute quality increases monotonically with budget.}
  \label{fig:token_scores}
\end{minipage}\hfill
\begin{minipage}[t]{0.48\textwidth}
  \centering
  \includegraphics[width=\linewidth]{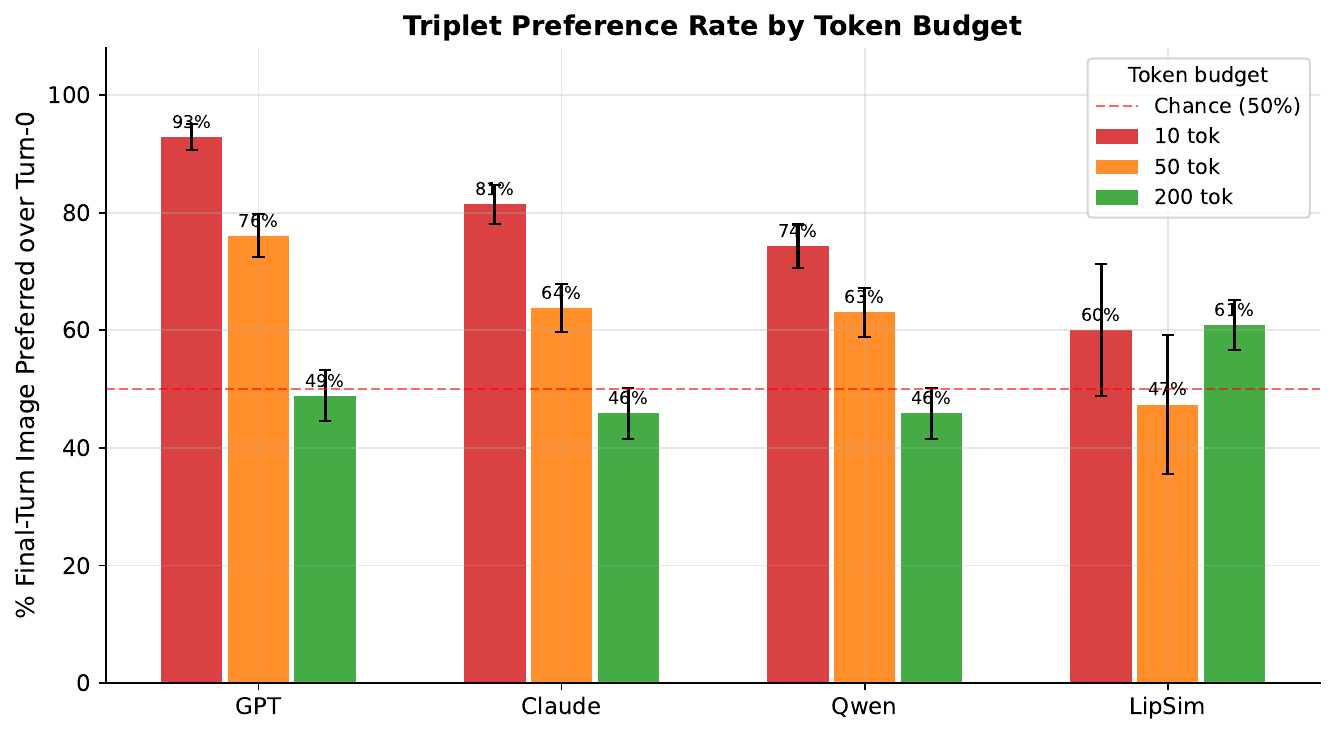}
  \caption{%
    Triplet preference rate for \textsc{gpt-5.2}~+~\textsc{gpt-image-1.5}
    at three token budgets, by judge and LipSim.
    Preference for the final-turn image decreases with budget; all conditions
    remain above the 50\% chance baseline (dashed line).}
  \label{fig:token_triplet}
\end{minipage}
\end{figure*}

Scores increase monotonically with budget (Figure~\ref{fig:token_scores}),
confirming that longer descriptions yield richer first renderings and higher
absolute quality at episode end. Triplet preference for the final-turn image
decreases monotonically (Figure~\ref{fig:token_triplet}): at 10~tokens the
gap between~$I_1$ and~$I_T$ is large and visually salient; at 200~tokens
$I_1$ is already detailed, leaving little room for visible improvement.

\subsection*{Score trajectories by category}

\begin{figure*}[p]
\centering
  \includegraphics[width=\textwidth]{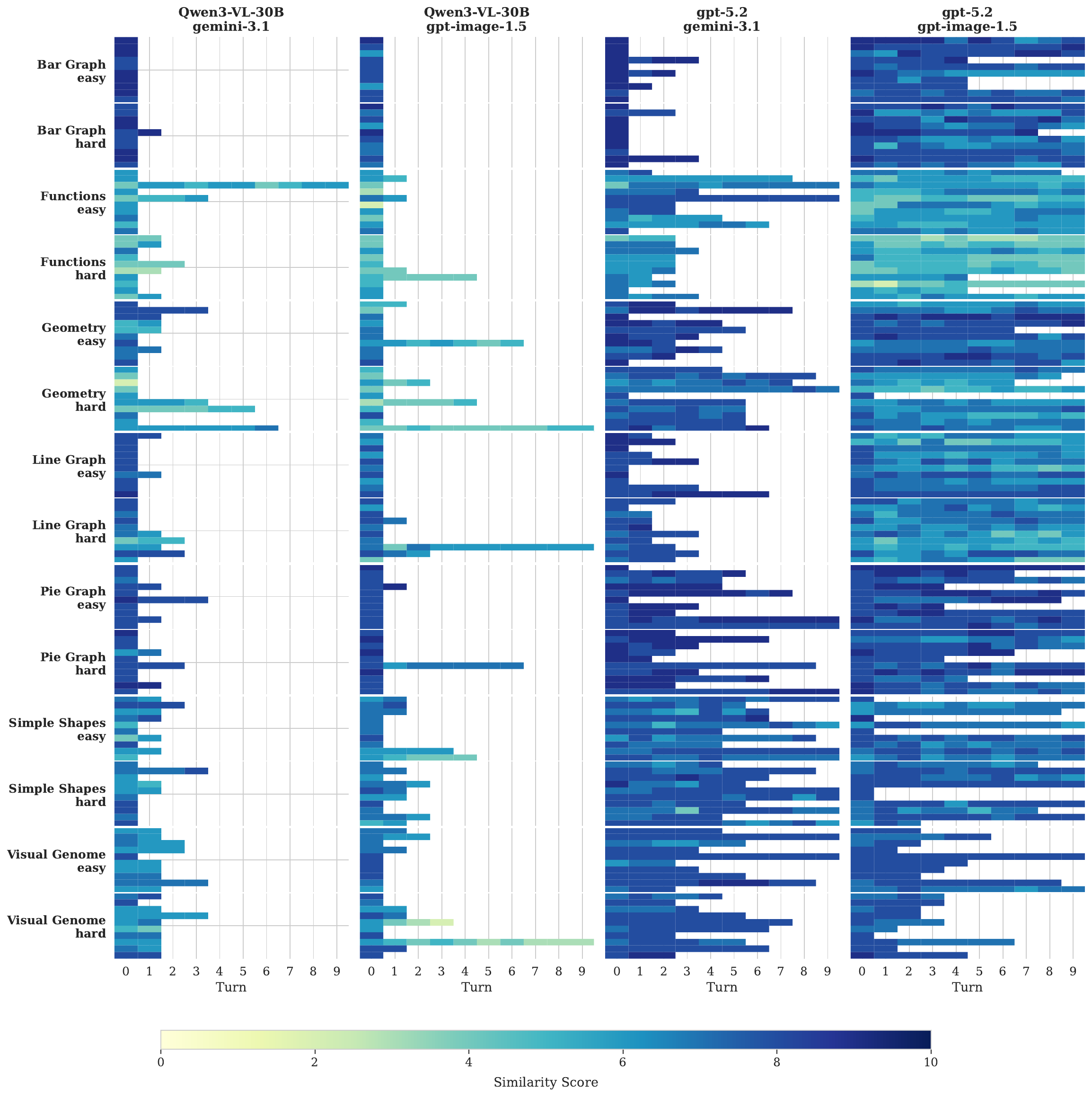}
  \caption{%
    Score trajectory heatmap: mean similarity score per turn, image category,
    and describer--generator pair.
    Rows are categories; columns are turns; colour encodes mean similarity. The turn scores are assigned by GPT-5.2.}
  \label{fig:trajectory_heatmap}
\end{figure*}

Figure~\ref{fig:trajectory_heatmap} shows mean similarity score per turn for
every category and model pair: \textsc{gpt-5.2} pairings improve steadily
across turns; \textsc{Qwen3-VL-30B} pairings plateau after turn~1.

\section{Prompt Templates}
\label{app:prompts}
The prompt templates are given in Figure~\ref{fig:init_prompt}, ~\ref{fig:init_prompt2}. The prompt template for generating synthetic images is given in Figure~\ref{fig:gen_syn_prompt}. We used \textit{GPT-5.2} for generating the synthetic images.

\begin{figure*}
  \centering
  \begin{prompt}
Let's play a game. I will feed you an image and you will produce
the most accurate description possible. The description should
capture every detail and be extremely precise. Your description must be at most 200 tokens long.\\ 

The output should consist of an initial tag <DESCRIPTION> followed by the description. \\

Reply just with the two tags and their related content.
  \end{prompt}

  \begin{prompt}
      You will now receive an image that was generated using your initial description. The goal is to generate an image as similar as possible to the original one. If you deem that the generated image is close enough to the original one, reply with "<DONE>". \\
      
      If not, you will reply with the changes needed to correct the image. The updated description should ONLY contain the changes/edits that have to be made to the previous image to make it look like the original image. Do NOT provide the full description again. Your description must be at most 200 tokens long. The output should consist of an initial tag <DESCRIPTION> ,followed by the changes or just a tag word <DONE> if you deem that the images are similar enough. Reply just with the two tags and the content related.\\
      
Then these steps will repeat until you find the generated image similar enough to the original image.\\

<GENERATED\_IMAGE>
  \end{prompt}

  \caption{Prompt templates for Describer player. The first template is for the initial turn and the next template is for consecutive turns.}
  \label{fig:init_prompt}
\end{figure*}

\begin{figure*}
  \centering
  \begin{prompt}
<DESCRIPTION WHAT TO GENERATE>
  \end{prompt}

  \begin{prompt}
      <DESCRIPTION WHAT TO GENERATE>\\

<PREVIOUS\_GENERATED\_IMAGE>
  \end{prompt}

  \caption{Prompt templates for Generator player. The first template is for the initial turn and the next template is for consecutive turns.}
  \label{fig:init_prompt2}
\end{figure*}

\begin{figure*}
  \centering
  \begin{prompt}
Generate a description for a simple geometry diagram.\\

Requirements:
\begin{itemize}
\item The diagram should be completely different from the ones generated early,vary the graphical settings and colours as well.
\item The diagram described should be similar to ones found in geometry book
\item The diagram should be simple and easily understandable
\end{itemize}

STRICT OUTPUT FORMAT REQUIREMENT (MANDATORY): \\
Output ONLY:\\

<DESCRIPTION> 
(technical construction details only)
</DESCRIPTION>\\
<SUMMARY>
(exactly two lines describing the resulting diagram)
</SUMMARY>\\
\end{prompt}
\begin{prompt}
When provided with an image description, generate the Python code required to visualize the image.
\\

Requirements:
\begin{itemize}

\item Output should consist only of the required code, placed between two <CODE> tags.
\item Do not display the plot using plt.show(); 
\item Do not include any title or legend in the picture.
\item !IMPORTANT! All figures should fit entirely in the image, adjust sizes if necessary
\item !IMPORTANT! THE CODE MUST CONSISTS OF A FUNCTION NAMED "GENERATE" THAT RETURNS THE PLOT OBJECT WHEN CALLED AND THAT TAKES NO INPUT
\item !IMPORTANT! code will be run using the exec command so make sure to include all required imports
\item !IMPORTANT! DO NOT use ax or subplots
\item !IMPORTANT: THE CODE MUST BE BETWEEN TWO <CODE TAGS> as in the following example \\

<CODE>
(code goes here)
</CODE>
\end{itemize}
  \end{prompt}
  \caption{Prompts used to generate synthetic data for the experiment using GPT-5.2. The first prompt produces the diagrams descriptions (here is shown an example for the easy geometry experiment), the second one produces the code to generate images with Matplotlib. The first prompt was sent to the LLM alongside with all the previously generated images summaries.}
  \label{fig:gen_syn_prompt}
\end{figure*}

\end{document}